\documentclass{article}
\usepackage{ijcai19}

\usepackage{times}
\usepackage{soul}
\usepackage{url}
\usepackage[hidelinks]{hyperref}
\usepackage[utf8]{inputenc}
\usepackage[small]{caption}
\usepackage{subcaption}
\usepackage{graphicx}
\usepackage{amssymb}
\usepackage{amsmath}
\usepackage{booktabs}
\usepackage{algorithm}
\usepackage{algorithmic}
\usepackage{multirow}
\usepackage{footnote}
\usepackage{pgffor}
\usepackage{subfiles}
\usepackage{array}
\usepackage[symbol]{footmisc}
\usepackage[toc,page]{appendix}

\usepackage[american]{babel}

\title{Inductive Guided Filter: Real-time Deep Image Matting with\\Weakly Annotated Masks on Mobile Devices}

\author{
Yaoyi Li$^1$\footnote{Work done as an intern at Versa}
\and
Jianfu Zhang$^1$\and
Weijie Zhao$^2$\And
Hongtao Lu$^1$
\affiliations
$^1$Shanghai Jiao Tong University\\
$^2$Versa
\emails
\{dsamuel, c.sis\}@sjtu.edu.cn,
weijie.zhao@versa-ai.com,
htlu@sjtu.edu.cn
}

\begin{document}

\maketitle

\begin{abstract}
  Recently, significant progress has been achieved in deep image matting. Most of the classical image matting methods are time-consuming and require an ideal trimap which is difficult to attain in practice. A high efficient image matting method based on a weakly annotated mask is in demand for mobile applications.
  In this paper, we propose a novel method based on 
  Deep Learning and Guided Filter, called Inductive Guided Filter, which can tackle the real-time general image matting task on mobile devices. We design a lightweight hourglass network to parameterize the original Guided Filter method that takes an image and a weakly annotated mask as input. Further, the use of Gabor loss is proposed for training networks for complicated textures in image matting. 
  Moreover, we create an image matting dataset MAT-2793 with 
  a variety of foreground objects. Experimental results demonstrate that our proposed method massively reduces running time with robust accuracy.
\end{abstract}

\section{Introduction}

Image matting is a fundamental problem in computer vision, which models an  image as a linear combination of a foreground and a background image:
\begin{equation}
I_i = \alpha_i F_i + (1-\alpha_i) B_i, \alpha_i \in [0, 1],
\end{equation}
where $ \alpha_i $ is the linear coefficient at a pixel position $ i $, $ F_i $ for the foreground pixel at $ i $ and $ B_i $ for the corresponding background pixel. Image matting task is more than a high-accurate segmentation and proposed for the natural image decomposition, which takes transparency into consideration. The generated alpha matte can highly reduce the workload and special requirement of image or video editing for advertisement, design, Vlog, film and so on. 
With rapid growth of the users who are using mobile devices to edit images or videos, a fast and accurate model which can enhance user experience is in high demand.

However, most of the classical methods \cite{levin2008closed,chen2013knn,cho2016natural,xu2017deep} are time-consuming which are not capable of running on mobile devices for real-time.

\begin{figure}[t]
    \centering
    \includegraphics[width = 0.8\columnwidth]{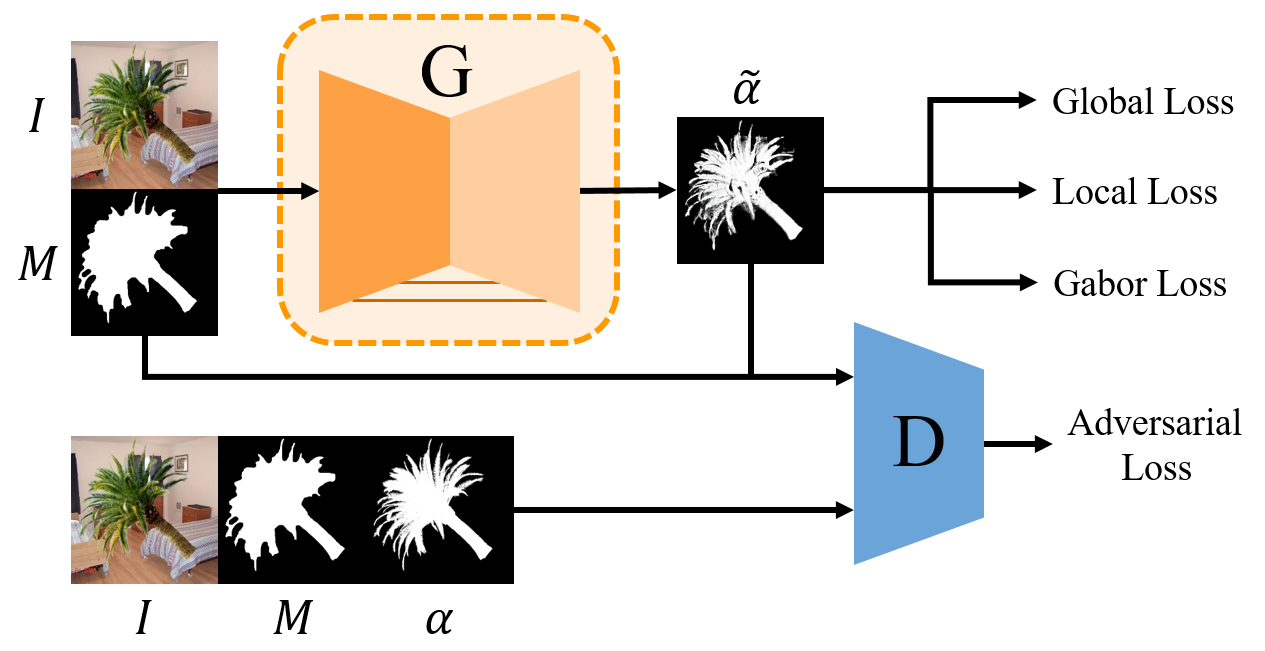}
    \caption{An illustration of our proposed method.}
    \label{fig:archtecture}
    \vspace{-5mm}
\end{figure}

Another obstacle for image matting on mobile devices is that the classical methods are sensitive with the input mask. Most of the time we can only obtain weakly annotated masks on mobile devices due to the limitation of time latency and computing ability. We coin the term ``weakly annotated mask'' to describe a noisy or inexact mask which gives some inaccurate annotations for foreground and background.
A weakly annotated mask can be an output binary image of a segmentation method, a thresholded depth map or an inexact annotated mask from user interaction. It contrasts conventional trimap which has an accurate annotation but consumes much time to compute. Reducing the quality of input masks or trimaps can massively degrade performance for the classical methods.

In this paper, we introduce a new deep learning framework that is specifically tailored for mobile devices by significantly reducing network parameters while retaining compatible accuracy. Compared with the classical methods that are highly dependent on the quality of trimap, our proposed model is robust with weakly annotated masks. We build our neural network for image matting as an image-to-image translation manner.
With the help of GAN framework and three other different losses, we can generate highly detailed image mattes with a tiny network.

\begin{figure*}[t]
    \centering
    \includegraphics[width = 0.8\textwidth]{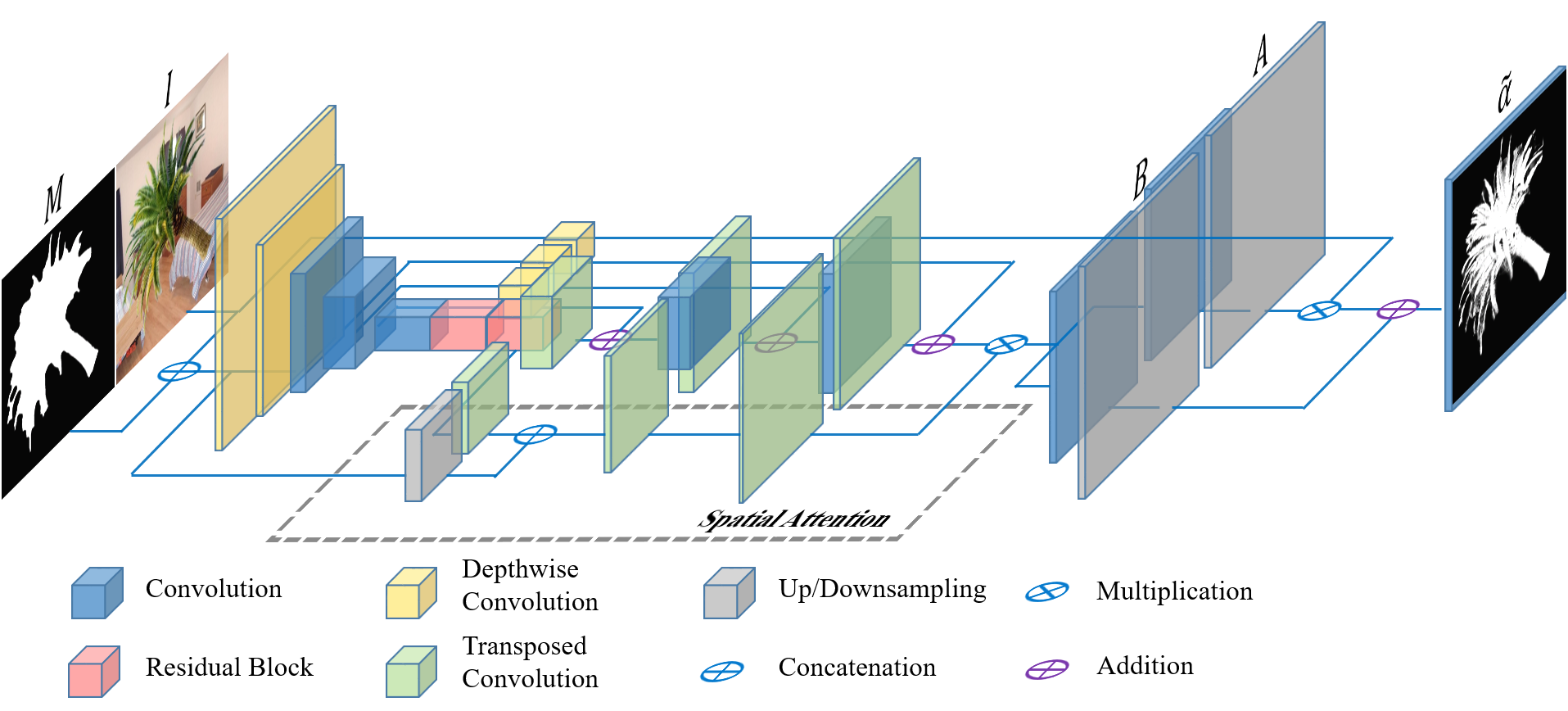}
    \caption[]{The overview of our generator. Linear coefficients $A$ and $B$ are generated from two branches sharing a same light weight Hourglass backbone. The right-most two upsampling layers both have a factor 4 which mimic Fast Guided Filter \cite{he2015fast} for acceleration.}
    \label{fig:generator}
    \vspace{-5mm}
\end{figure*}

Our main contributions in this paper are three-fold and can be summarized as followings:
\begin{itemize}
\item We design a novel real-time framework for the weakly annotated image matting task, dubbed Inductive Guided Filter. We are the first to introduce the combination of the deep convolutional neural networks and Guided Filter into the image matting task.
\item We propose a Gabor loss based on a bundle of Gabor filters to extract more comprehensive high-frequency features. To the best of our knowledge, no prior works have introduced such a loss function.
\item We further create an image matting dataset with 2793 foregrounds for the training of deep image matting called MAT-2793, which is the current biggest dataset for image matting to our knowledge. We evaluate our proposed method on MAT-2793 and Adobe Composition-1k testing dataset. Compared with the classical methods, our proposed method can achieve robust image matting effectively and efficiently.
\end{itemize}

\section{Related Works}

In this section, we review previous works on deep image matting and Guided Filter \cite{he2010guided} which are highly related to our method.
\paragraph{Deep Image Matting}

Many recent deep image matting approaches can be broadly categorized as general deep image matting approaches and ad hoc deep image matting approaches that are tailored for specific tasks. 

General deep image matting approaches attempt to predict the alpha matte given any natural image and the ideal trimap. Cho \textit{et al.} were the first to introduced deep learning into image matting task \shortcite{cho2016natural}. Their DCNN matting aimed to learn convolutional neural networks to combine the output of different classical image matting approaches. The Adobe Deep Image Matting (Adobe DIM) proposed in \cite{xu2017deep} was the first end-to-end deep image matting method, which significantly outperforms the conventional methods. Lutz \textit{et al.} further employed the generative adversarial networks (GAN) in their proposed AlphaGAN \cite{lutz2018alphagan}.

Some deep image matting methods are specialized in practical application scenarios like portrait matting. Shen \textit{et al.} proposed a portrait matting method \shortcite{shen2016deep}  with a deep network for trimap generation followed by a closed-form matting \cite{levin2008closed}, which can propagate gradients from closed-form matting to the neural network. Chen \textit{et al.} proposed a Semantic Human Matting \cite{chen2018semantic} which incorporated person segmentation and matting in an end-to-end manner. More lightweight deep image matting methods were proposed for portrait and hair matting on mobile devices \cite{zhu2017fast,levinshtein2018real,chen2019boundary}.

\paragraph{Guided Filter}

Guided Filter was proposed in \cite{he2010guided} as an edge-preserving smoothing operator that had a theoretical connection with the matting Laplacian matrix.
Deep Guided Filter \cite{wu2018fast} applied the Guided Filter to the output image of an image-to-image translation network as a super-resolution block and propagates the gradient through Guided Filter to the low-resolution output. Zhu \textit{et al.} proposed a fast portrait matting method with a feathering block inspired by Guided Filter \cite{zhu2017fast}, which will be elaborated in Section \ref{sec:igf}.

\section{Method}

\begin{figure*}[htp]
	\centering
	\begin{subfigure}{0.25\columnwidth}
		\centering
		{\includegraphics[width = 1\columnwidth]{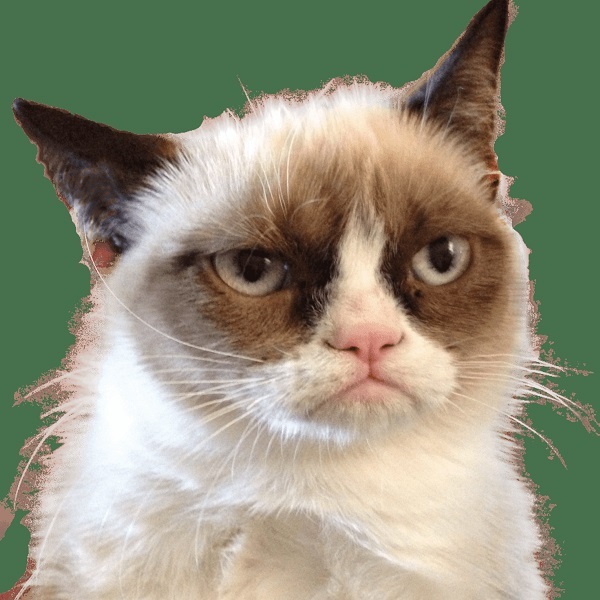}}
	\end{subfigure}	
	\begin{subfigure}{0.25\columnwidth}	
		\centering
		{\includegraphics[width = 1\columnwidth]{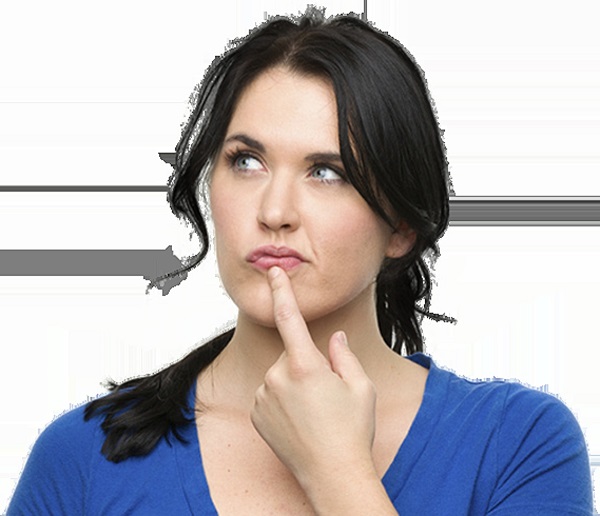}}
	\end{subfigure}	 
	\begin{subfigure}{0.25\columnwidth}	
		\centering
		{\includegraphics[width = 1\columnwidth]{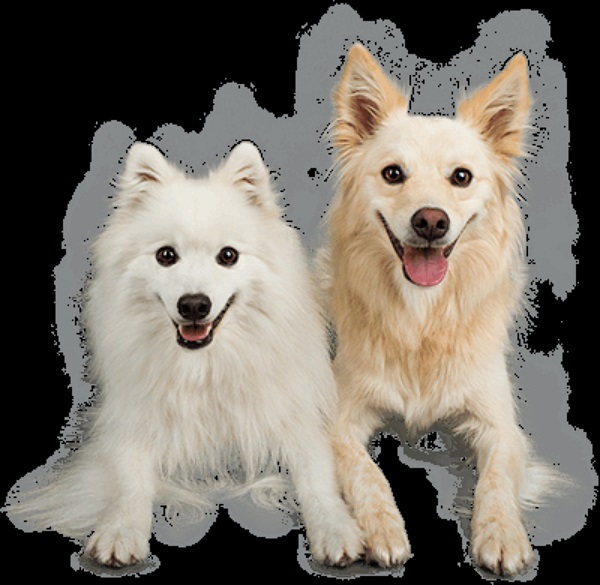}}
	\end{subfigure}	
	\begin{subfigure}{0.25\columnwidth}	
		\centering
		{\includegraphics[width = 1\columnwidth]{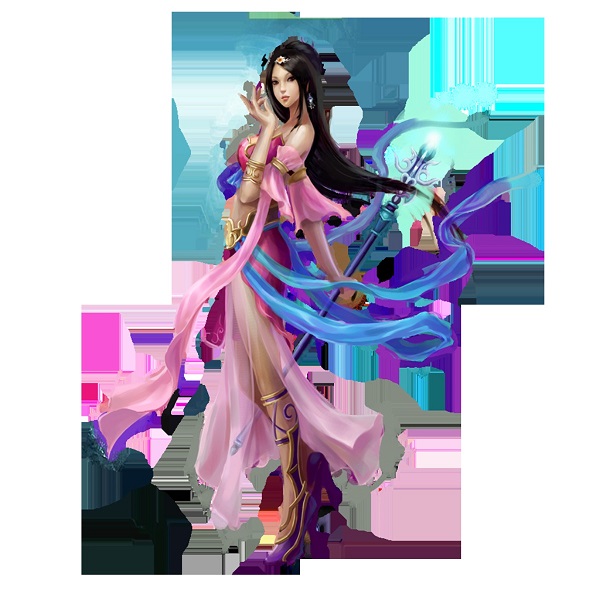}}
	\end{subfigure}	
	\begin{subfigure}{0.25\columnwidth}	
		\centering
		{\includegraphics[width = 1\columnwidth]{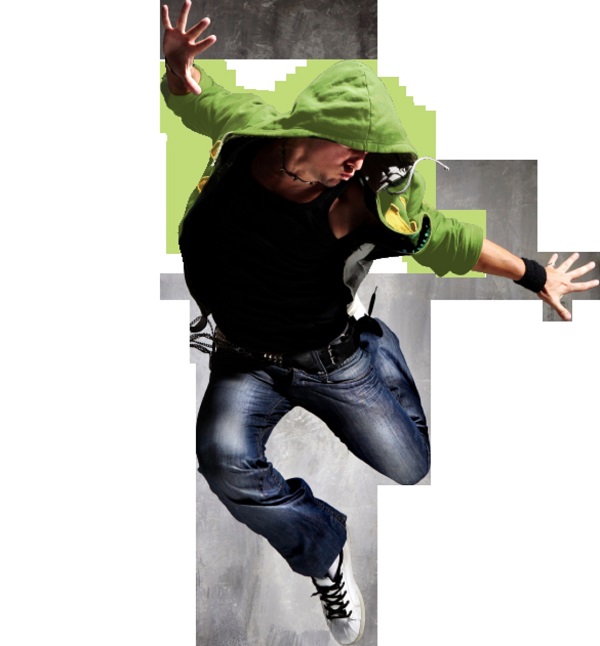}}
	\end{subfigure}		
	\begin{subfigure}{0.25\columnwidth}	
		\centering
		{\includegraphics[width = 1\columnwidth]{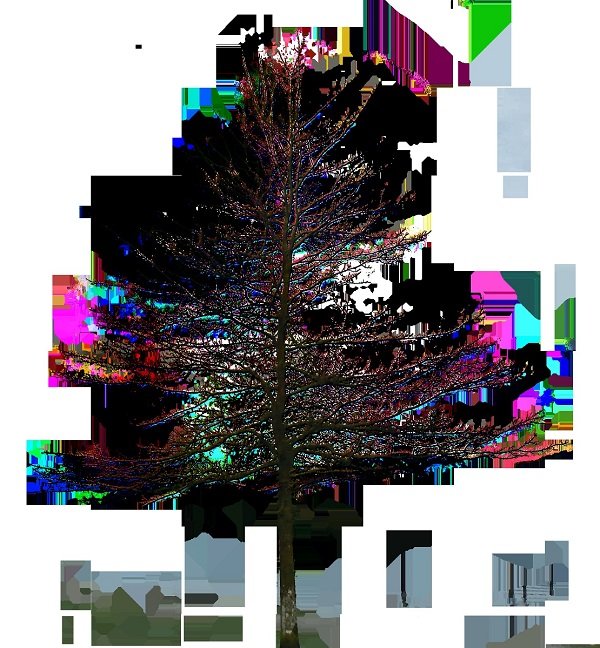}}
	\end{subfigure}	
	\begin{subfigure}{0.25\columnwidth}	
		\centering
		{\includegraphics[width = 1\columnwidth]{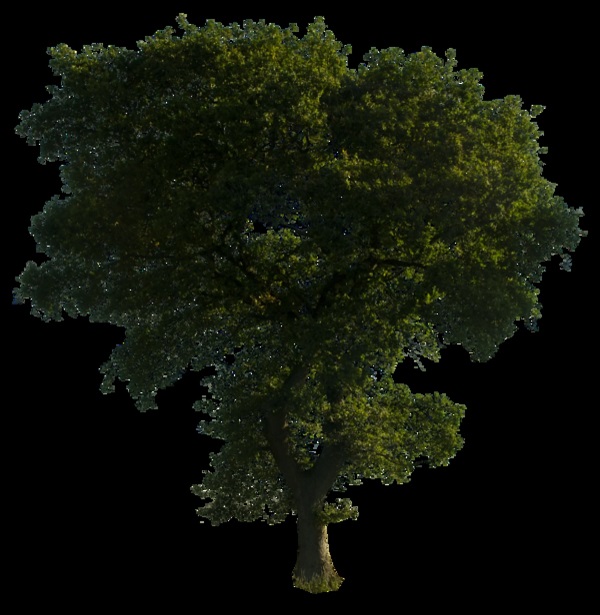}}
	\end{subfigure}	
	\begin{subfigure}{0.25\columnwidth}	
		\centering
		{\includegraphics[width = 1\columnwidth]{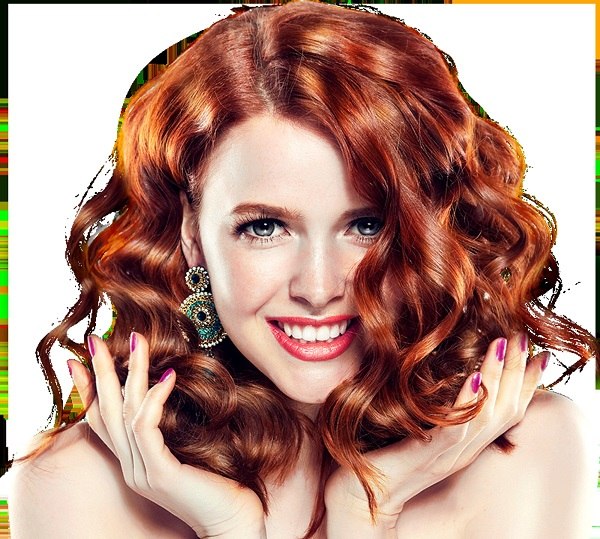}}
	\end{subfigure}	
	
	\begin{subfigure}{0.25\columnwidth}	
		\centering
		{\includegraphics[width = 1\columnwidth]{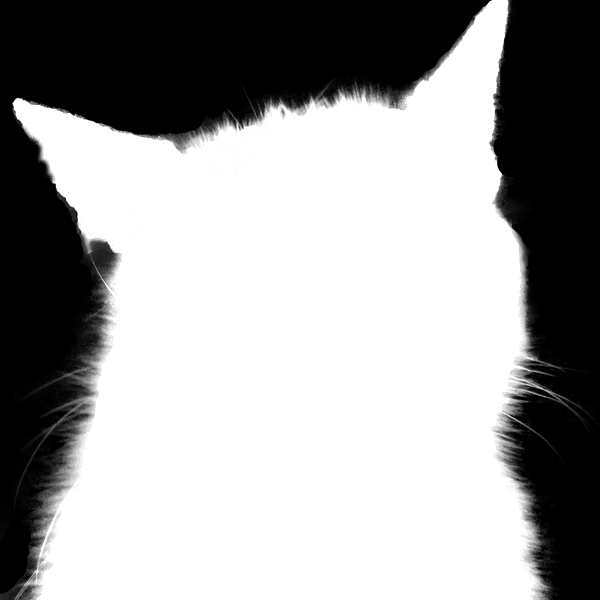}}
	\end{subfigure}	
	\begin{subfigure}{0.25\columnwidth}	
		\centering
		{\includegraphics[width = 1\columnwidth]{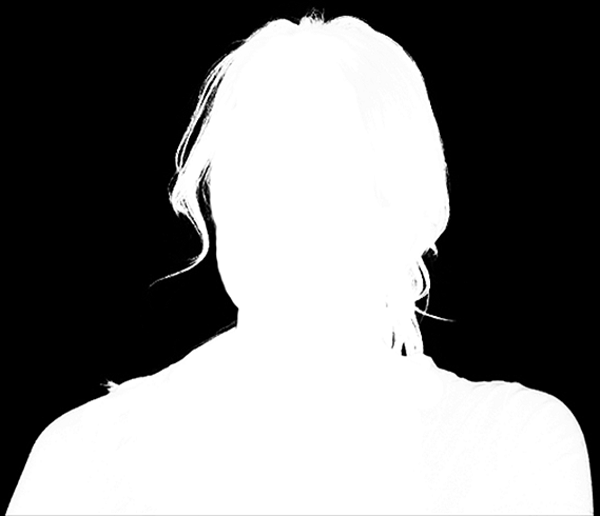}}
	\end{subfigure}	
	\begin{subfigure}{0.25\columnwidth}	
		\centering
		{\includegraphics[width = 1\columnwidth]{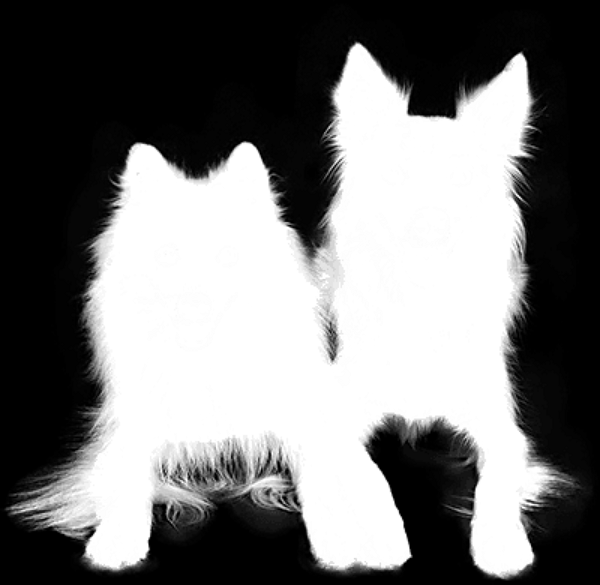}}
	\end{subfigure}	
	\begin{subfigure}{0.25\columnwidth}	
		\centering
		{\includegraphics[width = 1\columnwidth]{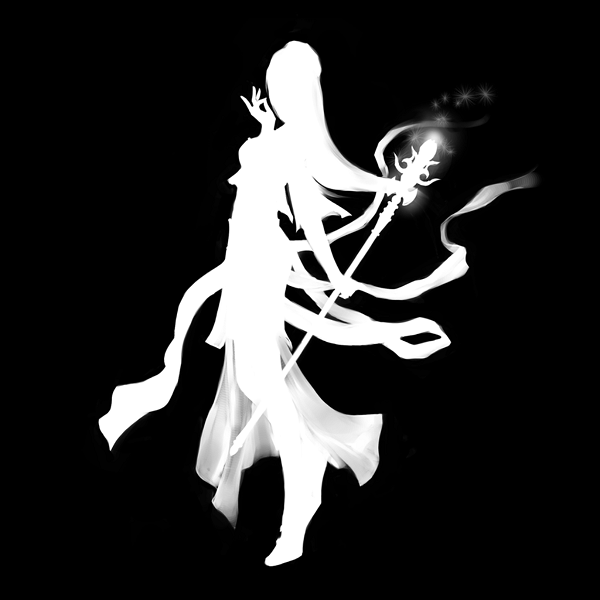}}
	\end{subfigure}	
	\begin{subfigure}{0.25\columnwidth}	
		\centering
		{\includegraphics[width = 1\columnwidth]{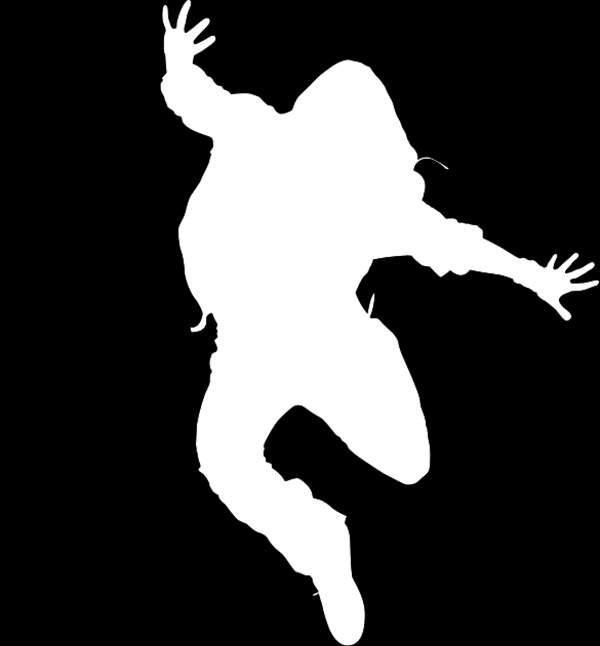}}
	\end{subfigure}		
	\begin{subfigure}{0.25\columnwidth}	
		\centering
		{\includegraphics[width = 1\columnwidth]{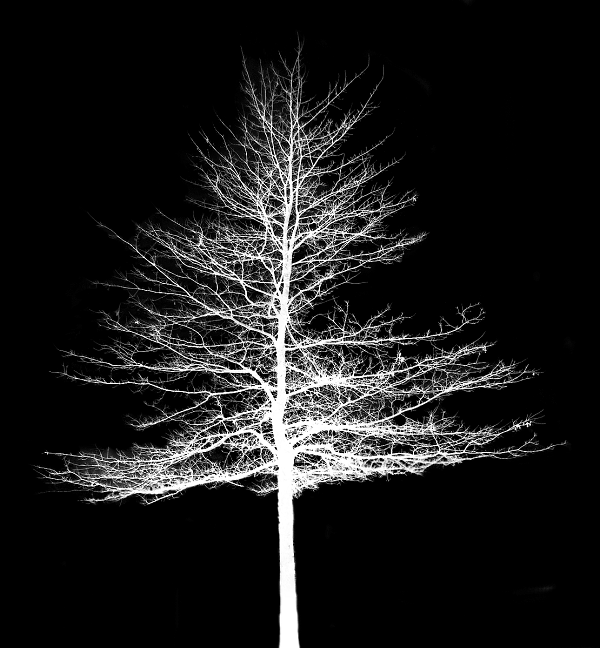}}
	\end{subfigure}	
	\begin{subfigure}{0.25\columnwidth}	
		\centering
		{\includegraphics[width = 1\columnwidth]{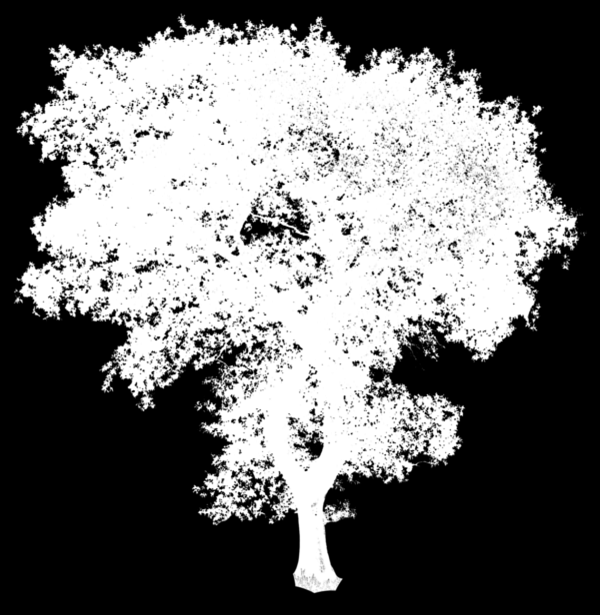}}
	\end{subfigure}	
	\begin{subfigure}{0.25\columnwidth}	
		\centering
		{\includegraphics[width = 1\columnwidth]{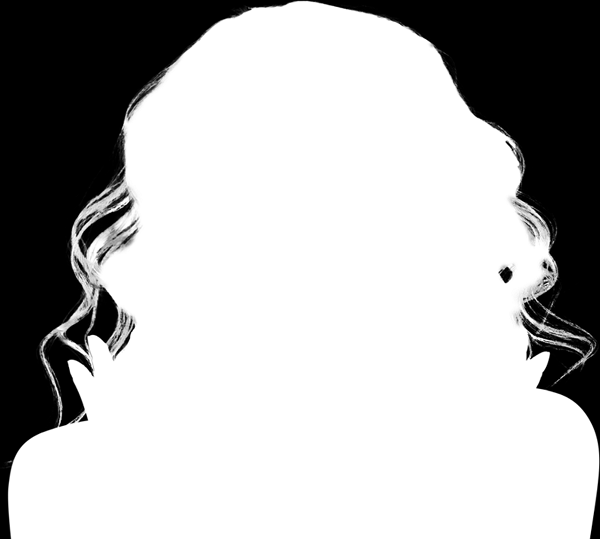}}
	\end{subfigure}	
	
	\caption{Some foreground object samples from the MAT-2793 dataset.}
	\label{fig:foreground}
	\vspace{-3mm}
\end{figure*}

We attempt to build an extremely high efficient image matting neural network which takes a weakly annotated mask as input. To this end, we employ the idea of linear model assumption in Guided Filter \cite{he2010guided}, which is robust and efficient in feathering tasks. 

Following \cite{lutz2018alphagan}, we adopt Generative Adversarial Network (GAN) \cite{goodfellow2014generative}  architecture to our model. The coarse architecture of our method is illustrated in Figure \ref{fig:archtecture} and details of generator in Figure \ref{fig:generator}. 

\subsection{Inductive Guided Filter Formulation}
\label{sec:igf}

In Guided Filter \cite{he2010guided}, the basic assumption is that the output alpha is a linear transform of
  guidance image $ I $ in a small window $ \omega_k $ centered at pixel $ k $:

\begin{equation}
\alpha_i =   A_{k}I_{i} + B_{k}, \forall i \in \omega_k,
\end{equation}
in which $ A_{k} $ and $ B_{k} $ are linear coefficients to be optimized. The optimization objective is to minimize the difference between output $ \alpha_i $ and the corresponding pixel $ M_{i} $ on input weakly annotated mask $ M $ with the regularization on $ A_{k} $. In the image matting setting, Guided Filter solves an optimization problem for each image and mask to generate a linear transformation from input image $ I $ to matte estimation $ \alpha $ which is as close to input mask $ M $ as possible.

Although Guided Filter is a fast and effective method for weakly annotated image matting task, it is limited by the constraint that the difference between optimal alpha matte and weakly annotated mask should be small enough. Empirically, the mask from a semantic segmentation method or user interaction will have a relatively large difference from the ground-truth alpha (see our testing set samples in Figure \ref{fig:test-data}).

Different from Guided Filter, our method attempts to deal with a supervised learning task instead of an optimization problem. We abandon the objective function and remove the constraint on the difference between matte estimation and mask. An inductive model based on the linear transform assumption is built to leverage the ground truth information in an image matting dataset. We formula the Inductive Guided Filter as 
\begin{equation}
\alpha = \phi_{A}(I, M)\circ I + \phi_{B}(I, M), 
\label{eq:igf}
\end{equation}
where $ \circ $ denotes Hadamard product and we parameterize $ A $ and $ B $ in Guided Filter by neural networks $ \phi_{A}(I, M) $ and $ \phi_{B}(I, M) $. Networks $ \phi_{A} $ and $ \phi_{B} $ take image $ I $ and weakly annotated mask $ M $ as input and share backbone parameters (see last two layers in Figure \ref{fig:generator} for details). The optimization objective of Inductive Guided Filter is to minimize the difference between the alpha matte prediction and ground truth.

For any image and mask, $ \phi_{A} $ and $ \phi_{B} $ can generate the specific coefficients $ A $ and $ B $, to build a linear transform model for alpha matte.

A similar idea was also mentioned in \cite{zhu2017fast}. The main difference between our method and theirs is that the authors of \cite{zhu2017fast} formulated their feathering block based on the closed-form solution of Guided Filter:
\begin{equation}
\alpha_i = A_k M^F_i + B_k M^B_i + C_k, \forall i \in \omega_k,
\label{eq:mobilematitng}
\end{equation}
where $ M^F $ and $M^B$ are masks for foreground and background. $ A_k, B_k $ and $ C_k $ are coefficients that parameterized by a neural network like $ \phi(\cdot) $ in our method. From Equation \eqref{eq:mobilematitng} we can derive that the output of their feathering block will only preserve the edge and gradient of the mask instead of the input image. It can be seen as an attention map on the mask. Consequently, a weakly annotated mask may lead to performance degradation.
On the contrary, Inductive Guided Filter can be regarded as an attention map on the original input image, which is the same as the linear transform in Guided Filter. It is more robust to the noise in a mask.

\begin{figure*}[th]
\foreach \t in {1,2,4,6}{%
\centering
\begin{subfigure}{0.24\columnwidth}	
	\centering
	{\includegraphics[width = 1\columnwidth]{figures/data/test_image_\t.jpg}}
\end{subfigure}	
\begin{subfigure}{0.24\columnwidth}
	\centering
	{\includegraphics[width = 1\columnwidth]{figures/data/test_alpha_\t.jpg}}
\end{subfigure}	
\begin{subfigure}{0.24\columnwidth}
	\centering
	{\includegraphics[width = 1\columnwidth]{figures/data/test_mask_\t.jpg}}
\end{subfigure}
\begin{subfigure}{0.24\columnwidth}
	\centering
	{\includegraphics[width = 1\columnwidth]{figures/data/test_trimap_\t.jpg}}
\end{subfigure}
\begin{subfigure}{0.24\columnwidth}
	\centering
	{\includegraphics[width = 1\columnwidth]{figures/data/test_gf_\t.jpg}}
\end{subfigure}
\begin{subfigure}{0.24\columnwidth}
	\centering
	{\includegraphics[width = 1\columnwidth]{figures/data/test_dim_\t.jpg}}
\end{subfigure}
\begin{subfigure}{0.24\columnwidth}
	\centering
	{\includegraphics[width = 1\columnwidth]{figures/data/test_idf_\t.jpg}}
\end{subfigure}
\begin{subfigure}{0.24\columnwidth}
	\centering
	{\includegraphics[width = 1\columnwidth]{figures/data/test_comp_\t.jpg}}
\end{subfigure}

}
\centering
\begin{subfigure}{0.24\columnwidth}	
	\centering
	{\includegraphics[width = 1\columnwidth]{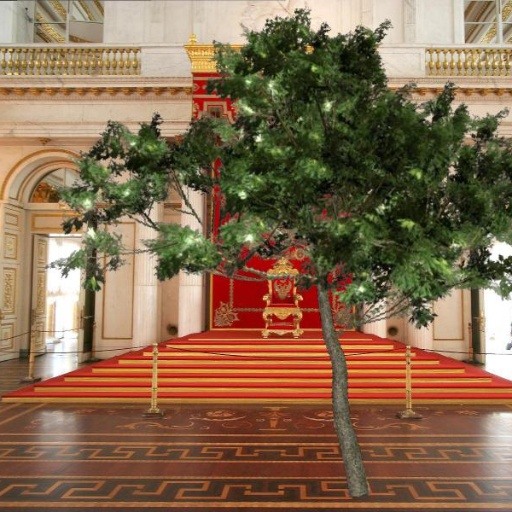}}
	\caption*{Image}
\end{subfigure}	
\begin{subfigure}{0.24\columnwidth}
	\centering
	{\includegraphics[width = 1\columnwidth]{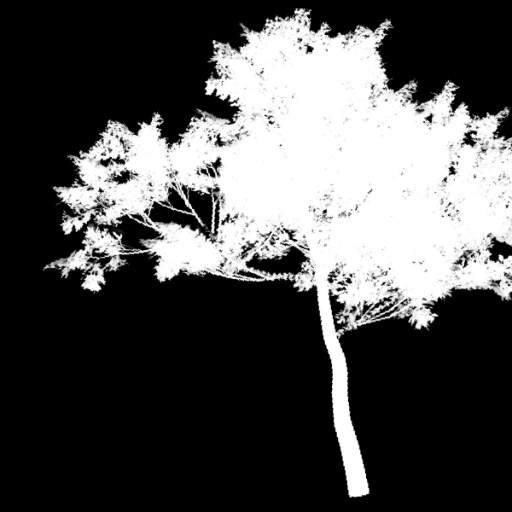}}
	\caption*{Ground Truth}
\end{subfigure}	
\begin{subfigure}{0.24\columnwidth}
	\centering
	{\includegraphics[width = 1\columnwidth]{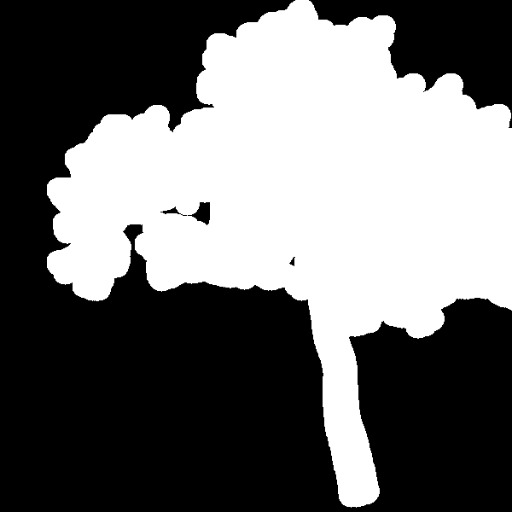}}
	\caption*{Mask}
\end{subfigure}
\begin{subfigure}{0.24\columnwidth}
	\centering
	{\includegraphics[width = 1\columnwidth]{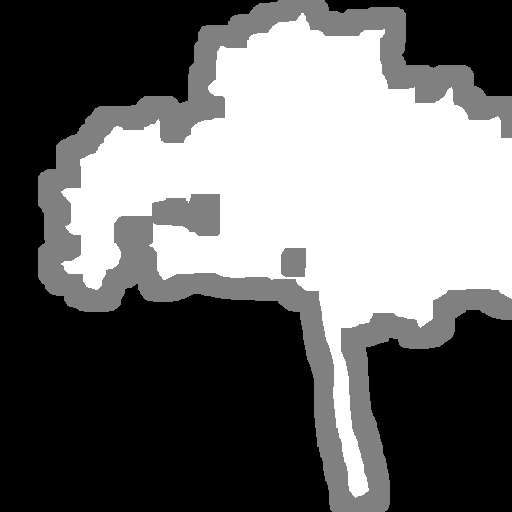}}
	\caption*{Trimap}
\end{subfigure}
\begin{subfigure}{0.24\columnwidth}
	\centering
	{\includegraphics[width = 1\columnwidth]{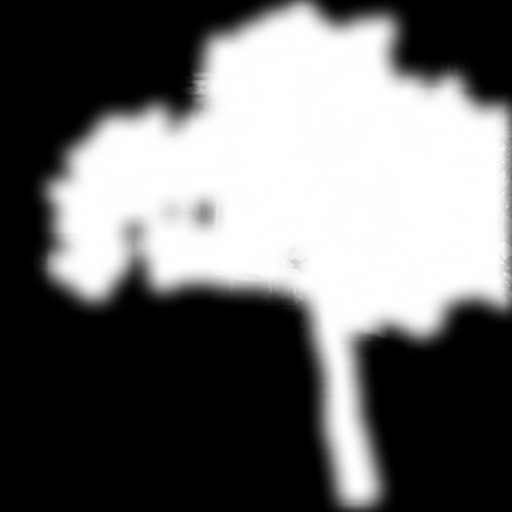}}
	\caption*{Guided Filter}
\end{subfigure}
\begin{subfigure}{0.24\columnwidth}
	\centering
	{\includegraphics[width = 1\columnwidth]{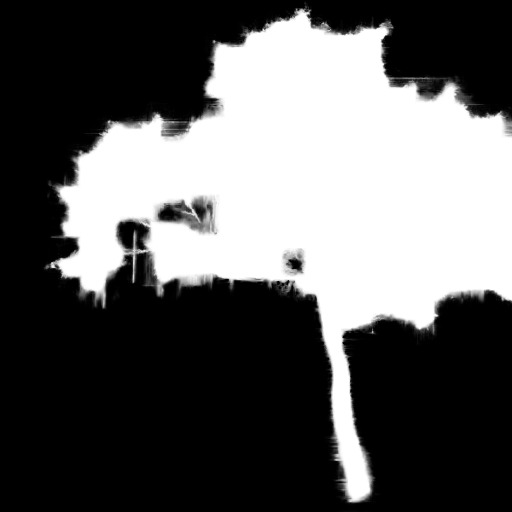}}
	\caption*{Adobe DIM}
\end{subfigure}
\begin{subfigure}{0.24\columnwidth}
	\centering
	{\includegraphics[width = 1\columnwidth]{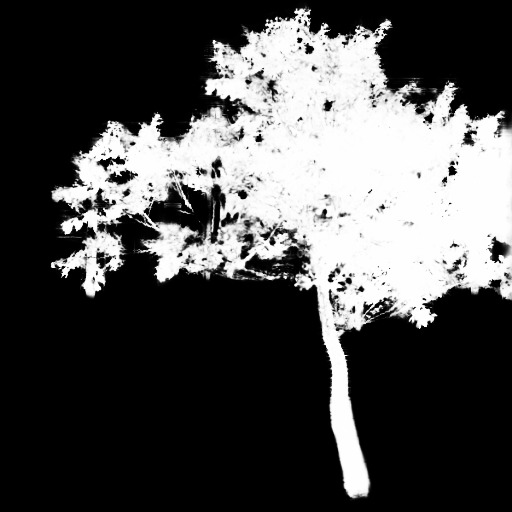}}
	\caption*{Ours}
\end{subfigure}
\begin{subfigure}{0.24\columnwidth}
	\centering
	{\includegraphics[width = 1\columnwidth]{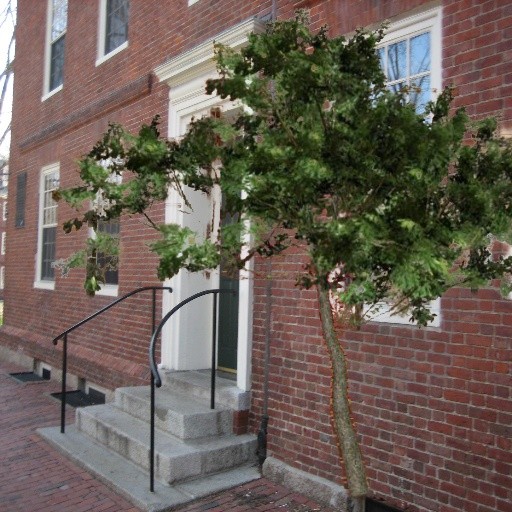}}
	\caption*{Composition}
\end{subfigure}

	\caption{The visual comparison results on MAT-2793 testing set. The trimap is only for Adobe DIM. The composition is composed with our results and random backgrounds.}
	\label{fig:test-data}
	\vspace{-3mm}
\end{figure*}

\subsection{Generator}

The generator consists of a lightweight Hourglass backbone, spatial attention mechanism, and a linear transformation. 

We build a lightweight Hourglass backbone following the structure of U-Net \cite{ronneberger2015u} and the Hourglass module in Stacked Hourglass Networks \cite{newell2016stacked} which prove to be effective to preserve low-level information from high-resolution features. Only two residual blocks
are involved in the bottleneck. Moreover, depthwise convolution, which is widely used in lightweight deep neural networks \cite{chollet2017xception,sandler2018mobilenetv2}, 
is adopted in the first two convolution layers and the shortcut connections. We only introduce depthwise blocks to the layers that have high-resolution feature maps as \cite{nekrasov2018real} did. There is no $1\times 1$ convolution layer between the first two depthwise convolutions. We regard them as adaptive downsamplings. All efforts aim to reduce inference latency. 

Spatial attention \cite{xu2015show,chen2017sca} has shown effectiveness in various computer vision tasks. Some previous deep image matting methods adopted spatial attention in their structures, which conduces to a decent matting performance \cite{chen2018semantic,zhu2017fast}. As for our attention mechanism, we fuse the feature from input and bottleneck to compute an attention map which is applied to the high-resolution features in the decoder.

Besides the adversarial loss, we impose three loss functions on the generator: global loss, local loss and Gabor loss. 
The real alpha matte and predicted alpha matte in triplet input are denoted by $ \alpha $ and $ \tilde{\alpha} $.

\paragraph{Global Loss} To supervise the alpha matte prediction, we leverage the global loss which is a L1 loss between the ground truth alpha and the estimated alpha, 
\begin{equation}
\mathcal{L}_g = \|\alpha-\tilde{\alpha}\|_1, 
\end{equation}
\paragraph{Local Loss} When we are training a network for image matting, in contrast to global loss, we would like objective function to focus more on the boundary of foreground object. Local loss is a weighted reconstruction based on a difference function $ \Delta(\alpha, M) = \delta(|\alpha - M|>\epsilon) $. The difference function yields a binary boundary map, in which $1$ for the same values in ground truth and mask and $0$ for the other pixels. $ \delta $ function enforces the small differences between below $ \epsilon $ are neglected. We use $ \epsilon=0.01 $ in the loss function. The local loss function is written as 
\begin{equation}
\mathcal{L}_l = \|\Delta(\alpha, M) \circ (\alpha-\tilde{\alpha})\|_1, 
\end{equation}
and in practice we apply an addition morphological dilation with a $ 7\times 7 $ kernel to the difference function for a larger boundary area.

\paragraph{Gabor Loss}

Perceptual loss proposed in \cite{johnson2016perceptual} dramatically improves the visual quality of predictions in supervised image transformation tasks. It provides a series of supervisions from high-frequency features to semantic level features. 
Perceptual loss utilizes a VGG network pretrained on RGB color images with specific classes. However, the alpha matte is a gray-scale image. Therefore, we design a Gabor loss to resemble the perceptual loss in our case. Gabor loss replaces the pretrained multi-layer kernels in perceptual loss with a set of single-layer Gabor filters to extract high-frequency features. 

Gabor filter was introduced into the neural network as a kernel or initialization in some previous works \cite{ouyang2013joint,luan2018gabor} due to its comparability to the kernels from shallow layers. Thus, we define the Gabor loss by
\begin{equation}
\mathcal{L}_{gb} = \sum_{\phi_{gb}\in \Phi}\|\phi_{gb}(\alpha)-\phi_{gb}(\tilde{\alpha})\|_2^2,
\end{equation}
where function $ \phi_{gb}(\cdot)$ denotes the convolution with Gabor filter, $ \Phi $ is the set of different Gabor filters. In our training, we design 16 different $7\times7$ Gabor filters with 16 orientations in $\Phi$. All of the filters have wavelength $\lambda=5$, spatial aspect ratio $\gamma=0.5$ and standard deviation $\sigma=0.5$ . We have also tried a larger set with different wavelengths and standard deviations, and no additional remarkable benefits were manifested in the training.

Some deep image matting method introduces gradient loss into their objective functions with a similar motivation \cite{levinshtein2018real,chen2019boundary}. Gradient loss minimizes the difference between the image gradient of the input image and predicted alpha, globally or locally. 
Comparing with image gradient, Gabor loss is an extended version with the capability to extract more comprehensive features, especially for alpha matte which is rich in the high-frequency component.

\subsection{Discriminator}

Lutz \textit{et al.} first introduced GAN into their proposed AlphaGAN \shortcite{lutz2018alphagan}. In AlphaGAN, the discriminator takes a trimap and a newly composited image from predicted alpha matte as its input. Since image matting does not focus on semantic information, it is ambiguous to judge whether a composited image is real or not. A composited image may have a high fidelity when it is generated from an incorrect or a partial alpha matte.

To overcome this, we feed the discriminator with a conditional triplet input which consists of an original image, a weakly annotated mask and an alpha matte, analogous to some methods with pair input \cite{chen2017multi,hu2018pose}. Given a triplet input, the discriminator can predict the self-consistency of an input. Concretely, the critic is designed to predict whether an estimated alpha matte is correct conditioned on the input image and mask.

\paragraph{Adversarial Loss}

We employ LSGAN \cite{mao2017least} with gradient penalty \cite{gulrajani2017improved} as the adversarial loss. 
The loss function is defined as 
\begin{equation}
\begin{aligned}
\mathcal{L}_{adv} =\; & \mathcal{L}_D + \mathcal{L}_G + \lambda_{gp}  \mathbb{E}_{\hat{\alpha}}[(\|\nabla_{\hat{\alpha}}D(\hat{\alpha}|I, M)\|_2-1)^2]\\
\mathcal{L}_D =\; & \mathbb{E}_\alpha[(D(\alpha|I, M)-1)^2] + \mathbb{E}_{\tilde{\alpha}}[D(\tilde{\alpha}|I, M)^2]\\
\mathcal{L}_G =\; & -\mathbb{E}_{\tilde{\alpha}}[(D(\tilde{\alpha}|I, M)-1)^2],
\end{aligned}
\end{equation}
where $ \hat{\alpha} $ is a convex combination of $ \alpha $ and $ \tilde{\alpha} $ with a random coefficient sampled from uniform distribution. We use $ \lambda_{gp} =10 $ in our training.

\begin{table}[t]
    \small
    \centering
    \begin{tabular}{lcccc}  
        \toprule
        Methods & MSE & SAD & Grad. ($ \times 10^3 $) &Conn. ($ \times 10^3 $)\\
        \midrule
        Guided Filter & 0.101 & 9.57 & 15.53 & 5.53\\
        Adobe DIM & 0.148 & 8.54 & 16.53 & 4.42\\
        Our method & \textbf{0.028} & \textbf{2.51} & \textbf{6.45} & \textbf{1.51}\\        
        \bottomrule
    \end{tabular}
\caption{The quantitative results on the MAT-2793 testing set}
\label{tab:ours}
\end{table}

\subsection{Full Loss and Implement Details}

The full loss function in Inductive Guided Filter is 
\begin{equation}
\mathcal{L} = \lambda_g \mathcal{L}_g + \lambda_l \mathcal{L}_l + \lambda_{gb} \mathcal{L}_{gb} + \lambda_{adv} \mathcal{L}_{adv},
\end{equation}
in which we use $ \lambda_g = 10, \lambda_l = 1, \lambda_{gb} = 200 $ and $ \lambda_{adv} = 1$.

We leverage PatchGAN \cite{isola2017image}, which is capable of discriminating the fidelity of local patches, to drive the attention of critic to detailed textures.
Spectral Normalization \cite{miyato2018spectral} has shown an appealing superiority in the training of GAN. We apply Spectral Normalization layers in our discriminator as well as the Batch Normalization \cite{ioffe2015batch}.

We incorporate training tricks: learning rate warm-up and cosine learning rate decay, following \cite{xie2018bag}. Adam optimizer \cite{kingma2014adam} is adopted with $ \beta_1=0.5 $, $ \beta_2=0.999$ and initial learning rate $ 0.0001 $ for both generator and discriminator.

The size of input images, masks, and output alpha matte are $512\times512$. 
Out-channels of the first 5 convolution layers in the generator are $ 4, 4, 8, 16, 32 $, and all $5$ convolution layers have a stride $2$. The whole network has $460,456$ trainable parameters with $400,990$ in discriminator and $59,466$ in the generator.

\begin{figure*}[th]
\foreach \t in {1,2,3}{%
\centering
\begin{subfigure}{0.24\columnwidth}	
	\centering
	{\includegraphics[width = 1\columnwidth]{figures/data/adobe_image_\t.jpg}}
\end{subfigure}	
\begin{subfigure}{0.24\columnwidth}
	\centering
	{\includegraphics[width = 1\columnwidth]{figures/data/adobe_gt_\t.jpg}}
\end{subfigure}	
\begin{subfigure}{0.24\columnwidth}
	\centering
	{\includegraphics[width = 1\columnwidth]{figures/data/adobe_mask_\t.jpg}}
\end{subfigure}
\begin{subfigure}{0.24\columnwidth}
	\centering
	{\includegraphics[width = 1\columnwidth]{figures/data/adobe_trimap_\t.jpg}}
\end{subfigure}
\begin{subfigure}{0.24\columnwidth}
	\centering
	{\includegraphics[width = 1\columnwidth]{figures/data/adobe_gf_\t.jpg}}
\end{subfigure}
\begin{subfigure}{0.24\columnwidth}
	\centering
	{\includegraphics[width = 1\columnwidth]{figures/data/adobe_dim_\t.jpg}}
\end{subfigure}
\begin{subfigure}{0.24\columnwidth}
	\centering
	{\includegraphics[width = 1\columnwidth]{figures/data/adobe_igf_\t.jpg}}
\end{subfigure}
\begin{subfigure}{0.24\columnwidth}
	\centering
	{\includegraphics[width = 1\columnwidth]{figures/data/adobe_comp_\t.jpg}}
\end{subfigure}

}
\centering
\begin{subfigure}{0.24\columnwidth}	
	\centering
	{\includegraphics[width = 1\columnwidth]{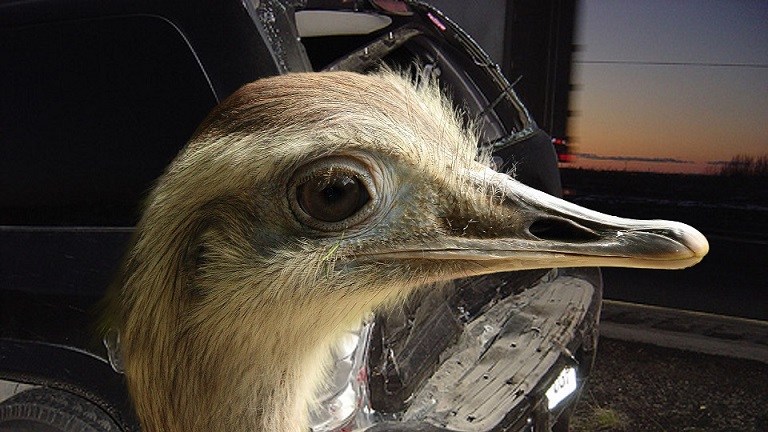}}
	\caption*{Image}
\end{subfigure}	
\begin{subfigure}{0.24\columnwidth}
	\centering
	{\includegraphics[width = 1\columnwidth]{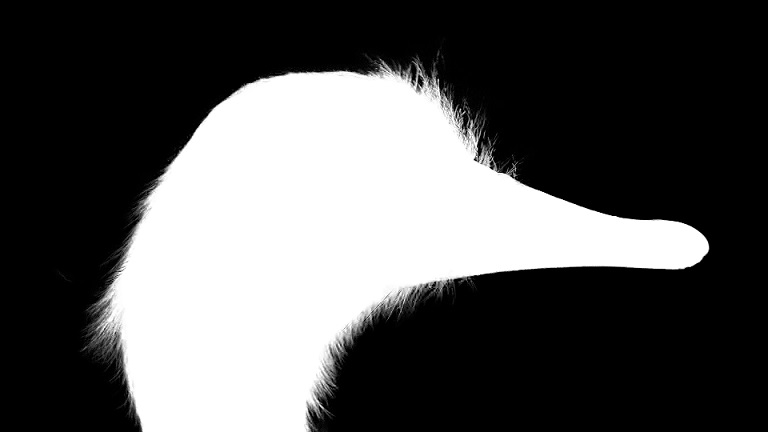}}
	\caption*{Ground Truth}
\end{subfigure}	
\begin{subfigure}{0.24\columnwidth}
	\centering
	{\includegraphics[width = 1\columnwidth]{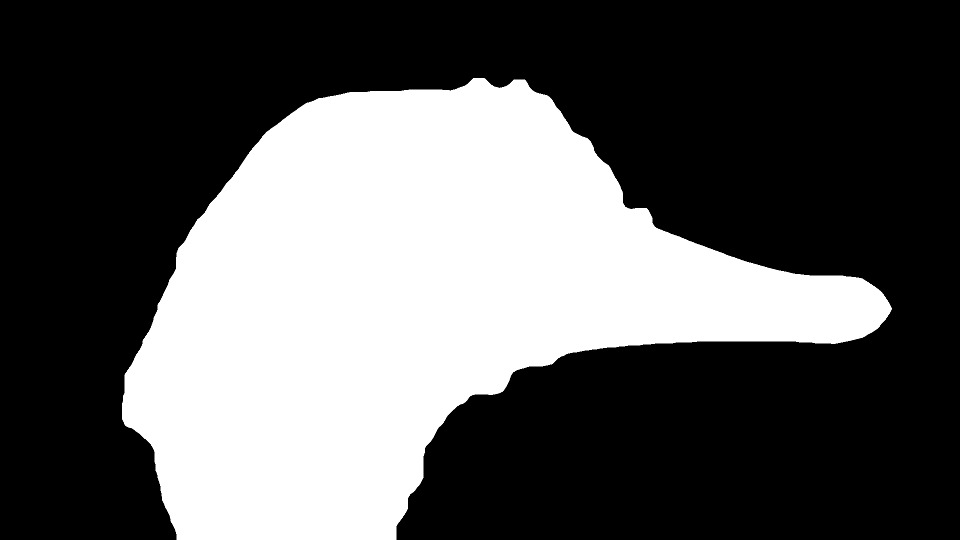}}
	\caption*{Mask}
\end{subfigure}
\begin{subfigure}{0.24\columnwidth}
	\centering
	{\includegraphics[width = 1\columnwidth]{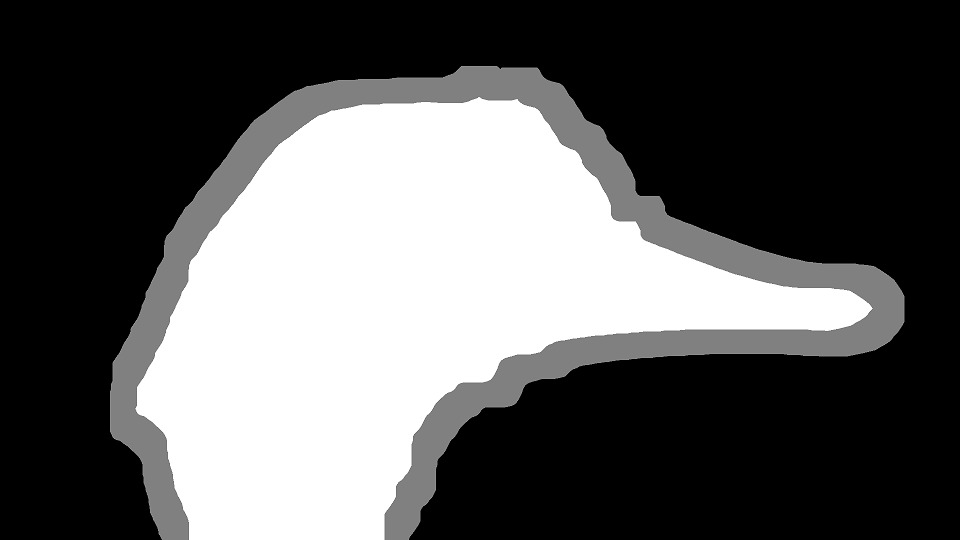}}
	\caption*{Trimap}
\end{subfigure}
\begin{subfigure}{0.24\columnwidth}
	\centering
	{\includegraphics[width = 1\columnwidth]{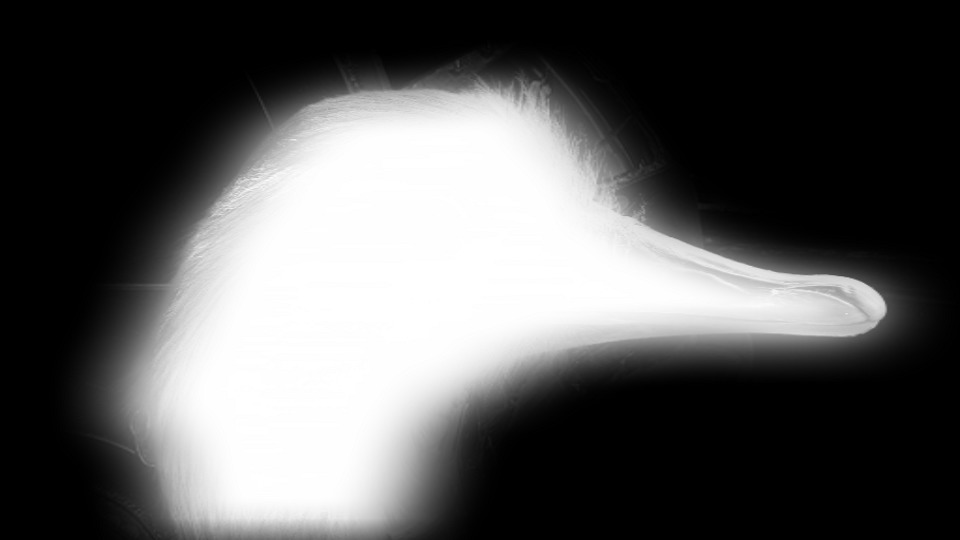}}
	\caption*{Guided Filter}
\end{subfigure}
\begin{subfigure}{0.24\columnwidth}
	\centering
	{\includegraphics[width = 1\columnwidth]{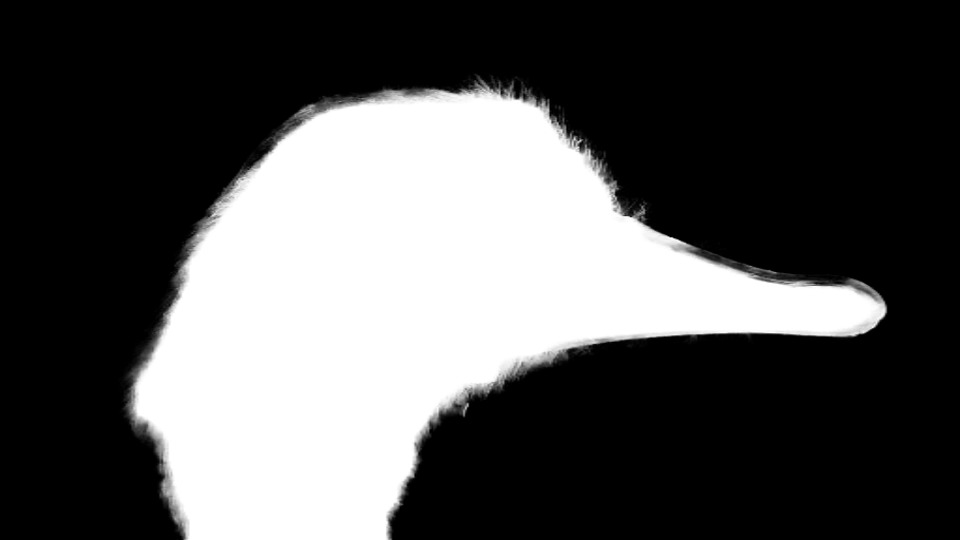}}
	\caption*{Adobe DIM}
\end{subfigure}
\begin{subfigure}{0.24\columnwidth}
	\centering
	{\includegraphics[width = 1\columnwidth]{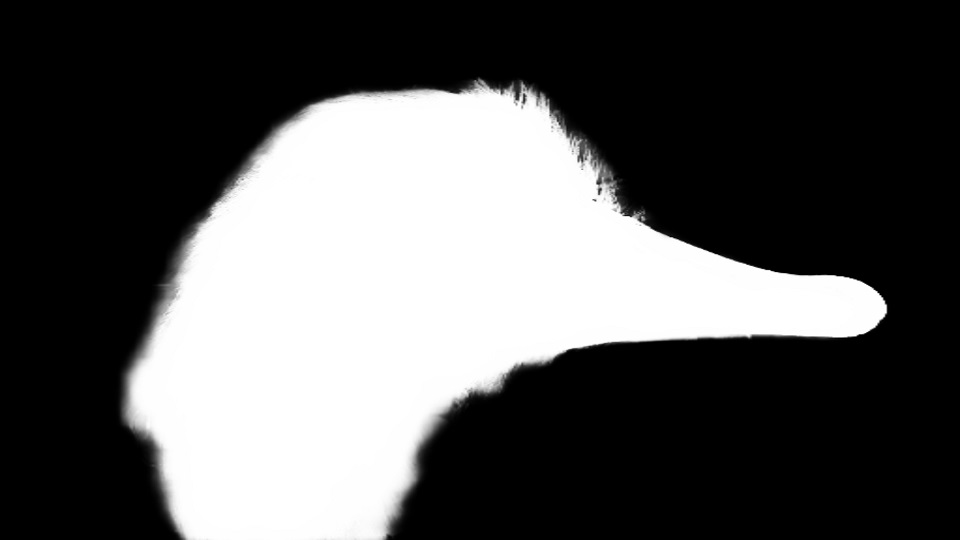}}
	\caption*{Ours}
\end{subfigure}
\begin{subfigure}{0.24\columnwidth}
	\centering
	{\includegraphics[width = 1\columnwidth]{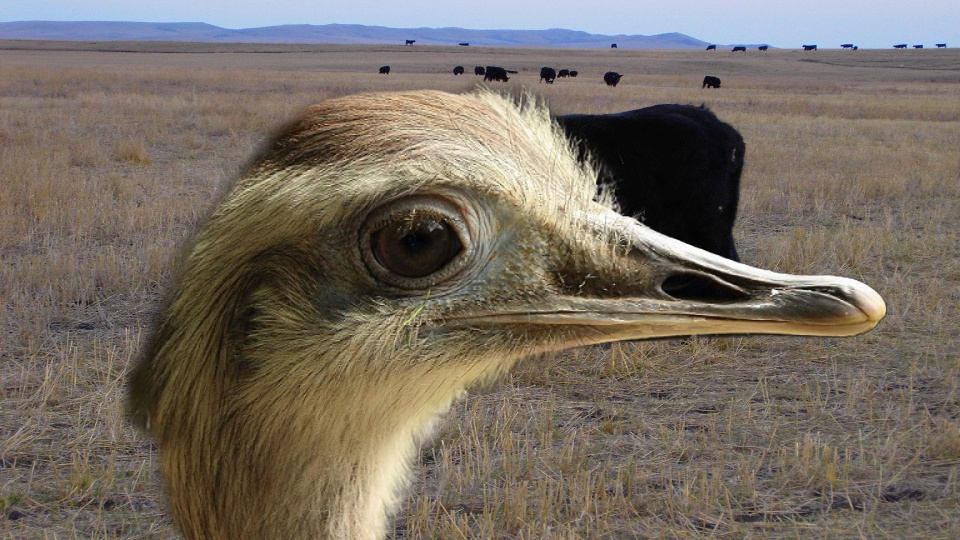}}
	\caption*{Composition}
\end{subfigure}

	\caption{The visual comparison results on Adobe Composition-1k. The trimap is only for Adobe DIM. The composition is composed with our results and random backgrounds.}
	\label{fig:adobe}
	\vspace{-3mm}
\end{figure*}

\section{Experiments}

In this section, we evaluate our method on two datasets, our MAT-2793 and Adobe Composition-1k \cite{xu2017deep}. We compare the proposed method with the state-of-the-art Adobe Deep Image Matting (Adobe DIM) \cite{xu2017deep} as well as Guided Filter \cite{he2010guided} quantitatively and qualitatively. Extensive experiments are conducted to make an efficiency comparison with some ad hoc real-time deep matting methods.


\begin{table}[t]
    \small
    \centering
    \begin{tabular}{lcccc}  
        \toprule
        Methods & MSE & SAD & Grad. ($ \times 10^3 $) &Conn. ($ \times 10^3 $)\\
        \midrule
        Guided Filter &  0.130 & 176.2  & 150.5  & 111.2 \\
        Adobe DIM & 0.243 & 105.2 & 55.0 & 56.1\\
        Our method & \textbf{0.064} & \textbf{46.2} & \textbf{42.1} & \textbf{35.7}\\        
        \bottomrule
    \end{tabular}
\caption{The quantitative results on the Adobe Composition-1k with weakly annotations.}
\label{tab:adobe}
\end{table}

\subsection{Dataset MAT-2793}
\label{sec:data}
We create an image matting dataset MAT-2793, which contains 2793 foreground objects, to tackle our weakly annotated image matting task. Most of the foregrounds and corresponding alpha mattes are gathered from the Internet as transparent PNG images especially from some free clipart website. Therefore, small parts of the foreground objects are not real-world images. We split our dataset into a training set with 2504 samples and a testing set with 289 samples.

In our experiments, we select 5360 high-resolution background images from the SUN dataset \cite{xiao2010sun} and ADE20k \cite{zhou2017scene}. We composite each object onto random backgrounds with a flipping and  11 different rotation angles to synthesis 22 different images. To generate weakly annotated masks, we first treat the alpha larger than $0.5$ as foreground. Then we apply dilation and erosion together to the foreground mask with random kernel size both from $5\times5$ to $30\times30$.
If the area of the foreground in a generated mask is less than half of the foreground in alpha matte, this training sample will be abandoned. We resize the shorter side of composited images, corresponding alpha mattes, and masks to 600 pixels. For data augmentation, we randomly crop the image and resize the crop to $512\times512$. Moreover, we randomly change the hue of images in the training phase. In testing, we make a $600\times600$ center crop and resize it to $512\times512$.


\subsection{Results on MAT-2793 testing set}
For a fair comparison between our method, Guided Filter and Adobe DIM\footnote[1]{We use the implementation from \url{https://github.com/foamliu/Deep-Image-Matting}}, we generate a trimap from each mask as the input of Adobe DIM. We apply a dilation along with an erosion on each mask both with a $20\times20$ kernel to create an unknown area. The errors are only computed in the unknown area. 
We implement the fast version of Guided Filter \cite{he2015fast} with downsampling factor 4 in all of our experiments.

We follow the image matting evaluation metrics suggested in \cite{rhemann2009perceptually}. We report the quantitative results under SAD, MSE, Gradient errors and Connectivity errors in Table \ref{tab:ours}. Furthermore, we display qualitative visual results in Figure \ref{tab:ours}. Results show that our proposed method is robust to noises in input masks and capable of capturing detailed texture in composited images.

\subsection{Results on Adobe Composition-1k}
Adobe Composition-1k is an image matting testing set with 1000 images and 50 unique foreground objects proposed in \cite{xu2017deep}.

In experiments on Adobe Composition-1k, we generate the weakly annotated mask in the same way as we did in Section \ref{sec:data}. To generate weakly trimaps for DIM, considering the large image size in Adobe Composition-1k, we apply dilation and erosion om masks with a $50\times50$ kernel. We then resize all the images to $512\times512$ in our experiments.


We illustrate the visual comparison results in Figure \ref{fig:adobe} and the quantitative results in Table \ref{tab:adobe}. All the errors are calculated in the unknown area of our generated weakly trimaps. The experiment results also show that our proposed method is capable of handling the weakly input.

\begin{savenotes}
\begin{table}
\small
\centering
\begin{tabular}{lccc}  
\toprule
Methods& Device & Batch Size & Time (ms)\\
\midrule
BANet-64\footnote[2]{Data from their original papers\label{ft:dagger}}  (512x512) & 1080 Ti& 1  & 23.3 \\
\midrule
LDN+FB\textsuperscript{\ref{ft:dagger}}  (128x128)& TITAN X & 1& 13  \\
\midrule
Adobe DIM (512x512)&  V100 & 1  & 51.1 \\
\midrule
\multirow{2}{*}{Ours w/ I/O (512x512)} & \multirow{2}{*}{ V100} &  1  & 3.48 \\
&&   256 & 1.46  \\
\midrule
\multirow{2}{*}{Ours w/o I/O (512x512)} & \multirow{2}{*}{ V100}  &  1 &  2.19  \\
& & 256  & 0.18 \\
\bottomrule
\end{tabular}
\caption{The results of speed evaluation on Nvidia GPU devices. We also display the speed of our method with or without GPU I/O.}
\label{tab:time_gpu}
\end{table}
\end{savenotes}

\subsection{Speed Evaluation}

We evaluate the efficiency of the proposed method on different platforms including Nvidia GPU, computer CPU, and mobile devices. Some real-time deep image matting methods for portrait (LDN+FB \cite{zhu2017fast}, BANet-64 \cite{chen2019boundary} or hair (HairMatteNet \cite{levinshtein2018real}) are also included in this evaluation.
We report the performance of these methods from the data in their original papers.

We demonstrate the inference speed on Nvidia GPU devices in Table \ref{tab:time_gpu} and speed on CPU or difference mobile devices in Table \ref{tab:time_mobie}. We deploy our model on different iPhone devices via the Apple CoreML framework. We can notice that taking a $512\times512$ image as input, our method can run at $5000+$ FPS on a single Nvidia V100 GPU with batch size 256 and achieve real-time performance on an iPhone SE in production in 2016.

\begin{savenotes}
\begin{table}
\centering
\begin{tabular}{lcc}  
\toprule
 Methods &   Device & Time (ms)   \\

\midrule
HairMatteNet\textsuperscript{\ref{ft:dagger}} (224x224)& iPad Pro GPU & 30   \\
\midrule
\multirow{2}{*}{LDN+FB\textsuperscript{\ref{ft:dagger}}(128x128)}  &  Core E5-2660 & 38  \\
 &  Adreno 530 & 62  \\
\midrule
Guided Filter (512x512)& Core i7-7700& 62.0  \\
\midrule
Adobe DIM (512x512) & Core i7-7700 & 4003.2  \\
\midrule
\multirow{4}{*}{Ours (512x512)}& Core i7-7700 & 13.0  \\
& iPhone Xs   & 15.7  \\
& iPhone X    & 22.3   \\
& iPhone SE  & 25.9   \\
\bottomrule
\end{tabular}
\caption{The results of speed evaluation on CPU and mobile devices.}
\label{tab:time_mobie}
\end{table}
\end{savenotes}

\section{Conclusion}
In this paper, we propose a extremely efficient method for weakly annotated image matting on mobile devices, dubbed Inductive Guided Filter. A lightweight hourglass backbone and a novel Gabor loss are leveraged in the model. We also create a large image matting  dataset MAT-2793. Evaluation on two testing datasets demonstrates that our proposed model is robust to the weakly annotated input mask and is competent to extract texture details in an image matting task.

{\small
\bibliographystyle{named}
\bibliography{main}}

\begin{appendices}



\section{Results of Real Images}
\begin{figure*}[th]
	\foreach \t in {1,...,5}{%
		\centering
		\begin{subfigure}{0.4\columnwidth}	
			\centering
			{\includegraphics[width = 1\columnwidth]{figures/supp/plant/merged/\t.jpg}}
		\end{subfigure}	
		\begin{subfigure}{0.4\columnwidth}
			\centering
			{\includegraphics[width = 1\columnwidth]{figures/supp/plant/mask/\t.png}}
		\end{subfigure}
		\begin{subfigure}{0.4\columnwidth}
			\centering
			{\includegraphics[width = 1\columnwidth]{figures/supp/plant/alpha/\t.jpg}}
		\end{subfigure}
		
	}
	\centering
	\begin{subfigure}{0.4\columnwidth}	
		\centering
		{\includegraphics[width = 1\columnwidth]{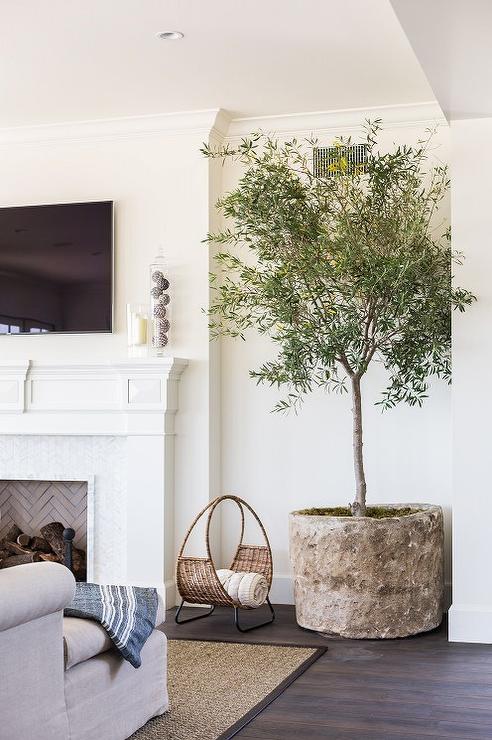}}
		\caption*{Image}
	\end{subfigure}	
	\begin{subfigure}{0.4\columnwidth}
		\centering
		{\includegraphics[width = 1\columnwidth]{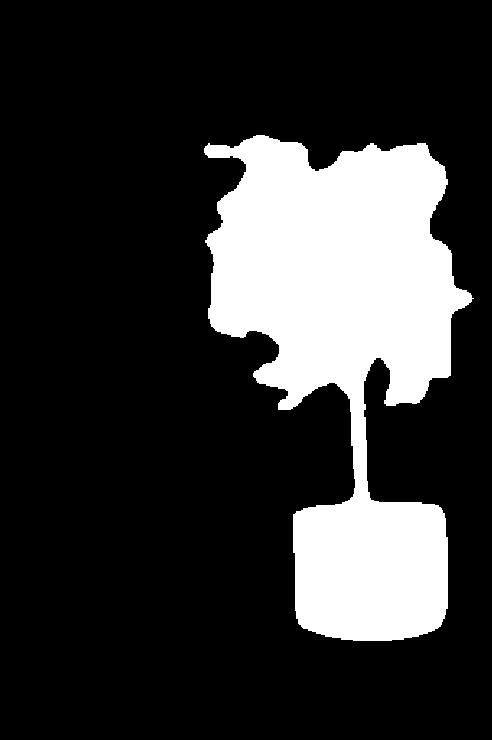}}
		\caption*{DeepLab v3+}
	\end{subfigure}
	\begin{subfigure}{0.4\columnwidth}
		\centering
		{\includegraphics[width = 1\columnwidth]{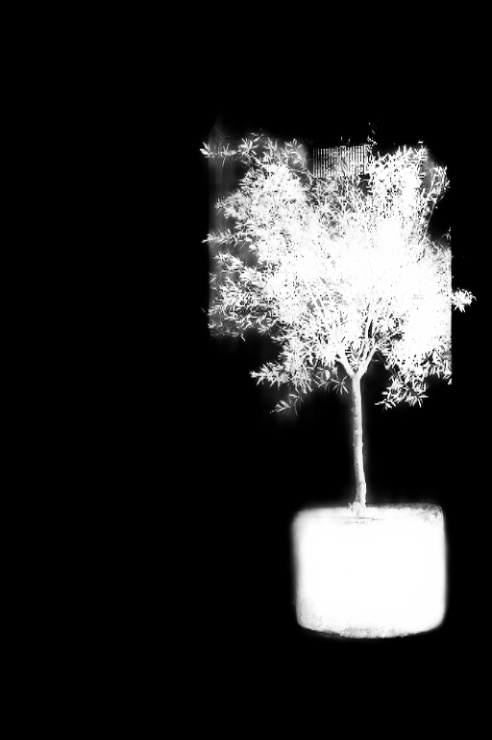}}
		\caption*{Ours}
	\end{subfigure}
	
	\caption{Results of some potted plant images. Masks are generated by DeepLab v3+.}
	\label{fig:plant_deeplab}
	\vspace{-3mm}
\end{figure*}

\begin{figure*}[th]
\foreach \t in {1,...,5}{%
	\centering
	\begin{subfigure}{0.5\columnwidth}	
		\centering
		{\includegraphics[width = 1\columnwidth]{figures/supp/supervisely/merged/\t.jpg}}
	\end{subfigure}	
	\begin{subfigure}{0.5\columnwidth}
		\centering
		{\includegraphics[width = 1\columnwidth]{figures/supp/supervisely/mask/\t.png}}
	\end{subfigure}
	\begin{subfigure}{0.5\columnwidth}
		\centering
		{\includegraphics[width = 1\columnwidth]{figures/supp/supervisely/alpha/\t.jpg}}
	\end{subfigure}
	
}
\centering
\begin{subfigure}{0.5\columnwidth}	
	\centering
	{\includegraphics[width = 1\columnwidth]{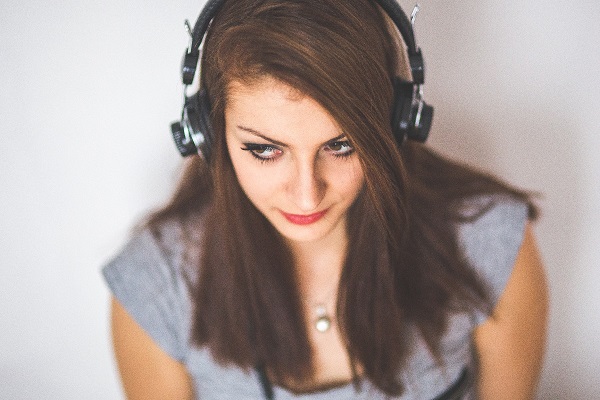}}
	\caption*{Image}
\end{subfigure}	
\begin{subfigure}{0.5\columnwidth}
	\centering
	{\includegraphics[width = 1\columnwidth]{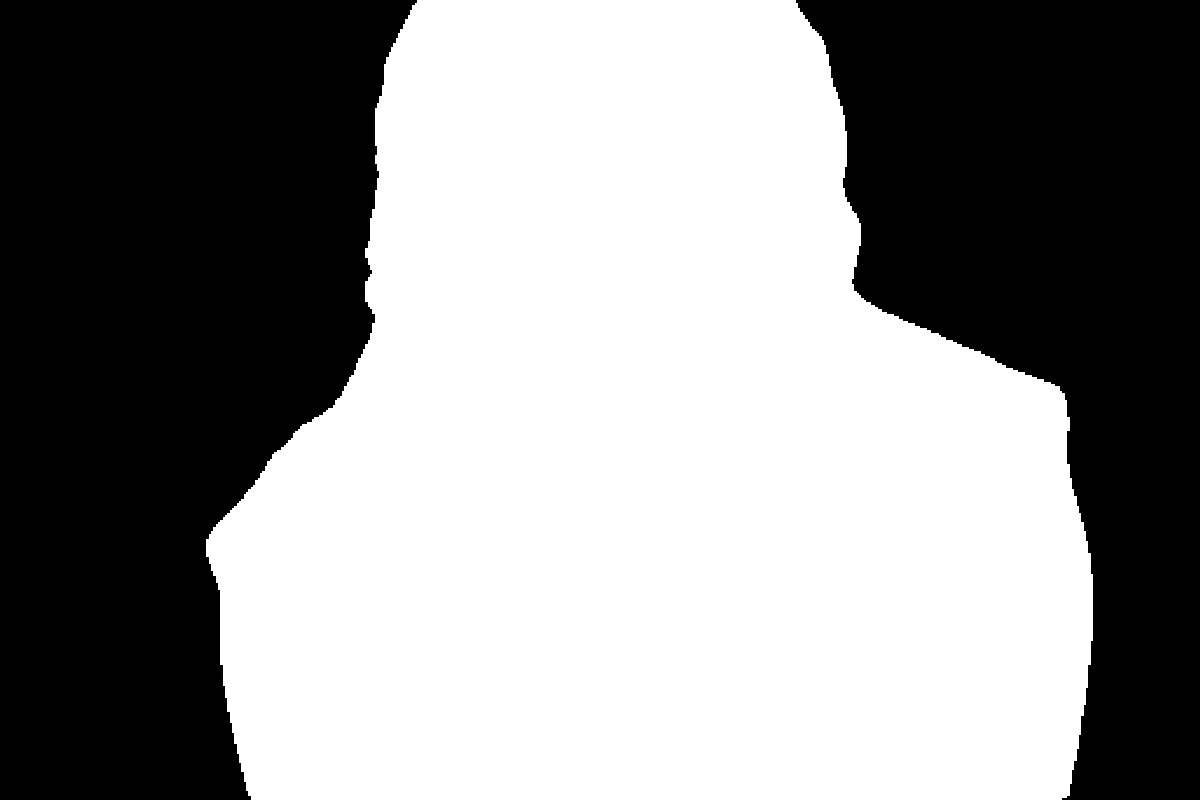}}
	\caption*{DeepLab v3+}
\end{subfigure}
\begin{subfigure}{0.5\columnwidth}
	\centering
	{\includegraphics[width = 1\columnwidth]{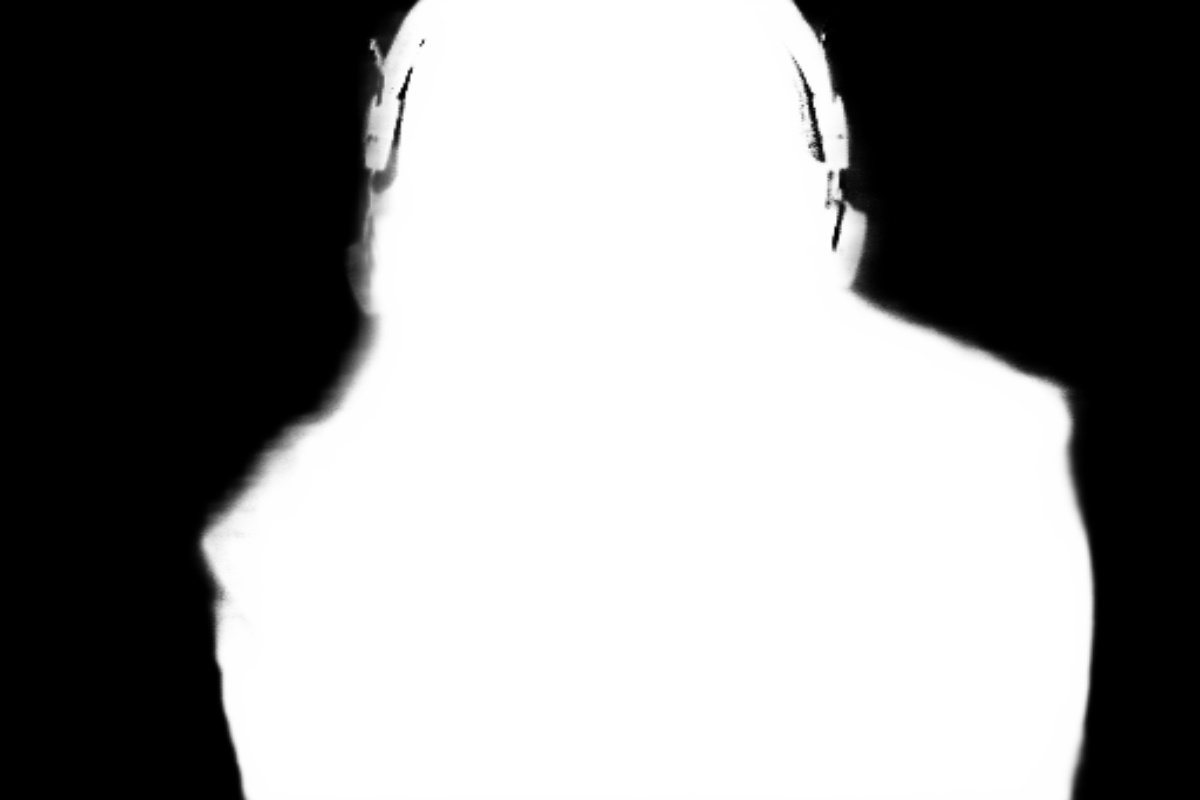}}
	\caption*{Ours}
\end{subfigure}

\caption{Results of samples from Supervisely Person dataset. Masks are generated by DeepLab v3+.}
\label{fig:sup_deeplab}
\vspace{-3mm}
\end{figure*}

\begin{figure*}[th]
	\foreach \t in {1,4,3,2}{%
		\centering
		\begin{subfigure}{0.5\columnwidth}	
			\centering
			{\includegraphics[width = 1\columnwidth]{figures/supp/adobe_deeplab/merged/\t.jpg}}
		\end{subfigure}	
		\begin{subfigure}{0.5\columnwidth}
			\centering
			{\includegraphics[width = 1\columnwidth]{figures/supp/adobe_deeplab/mask/\t.png}}
		\end{subfigure}
		\begin{subfigure}{0.5\columnwidth}
			\centering
			{\includegraphics[width = 1\columnwidth]{figures/supp/adobe_deeplab/alpha/\t.png}}
		\end{subfigure}
		
	}
	\centering
	\begin{subfigure}{0.5\columnwidth}	
		\centering
		{\includegraphics[width = 1\columnwidth]{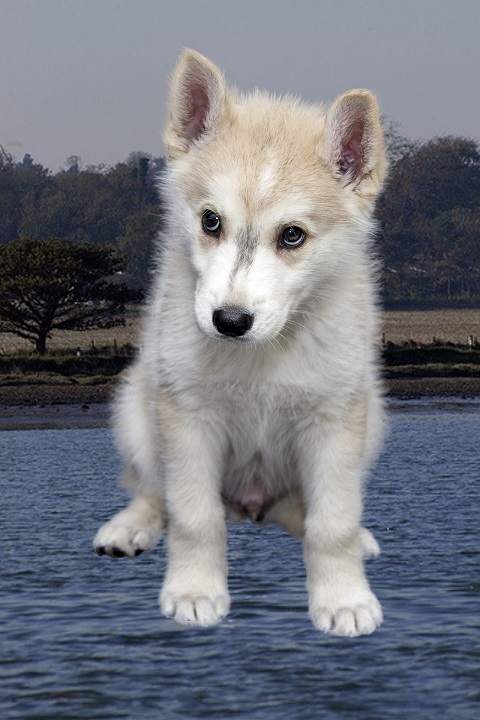}}
		\caption*{Image}
	\end{subfigure}	
	\begin{subfigure}{0.5\columnwidth}
		\centering
		{\includegraphics[width = 1\columnwidth]{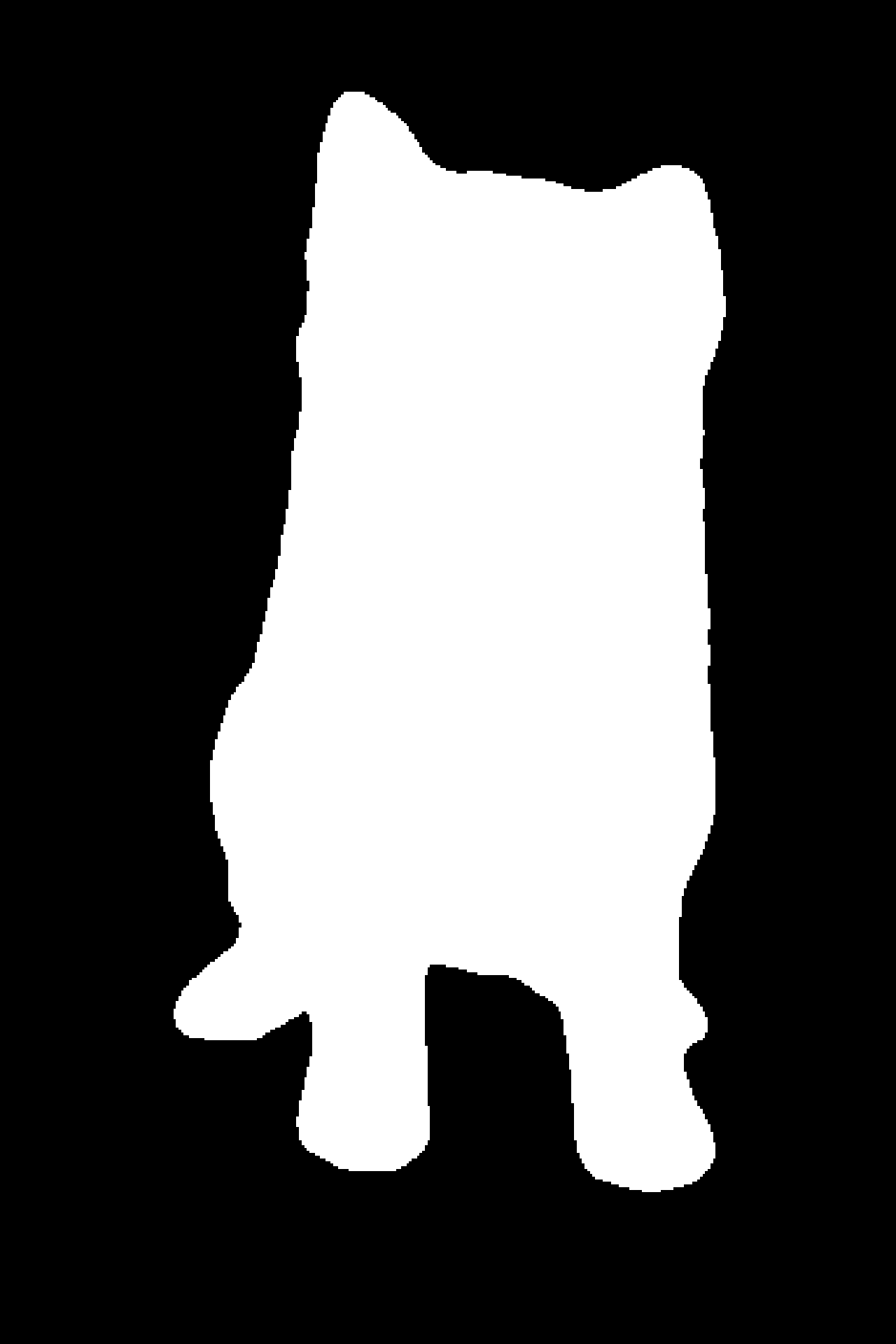}}
		\caption*{DeepLab v3+}
	\end{subfigure}
	\begin{subfigure}{0.5\columnwidth}
		\centering
		{\includegraphics[width = 1\columnwidth]{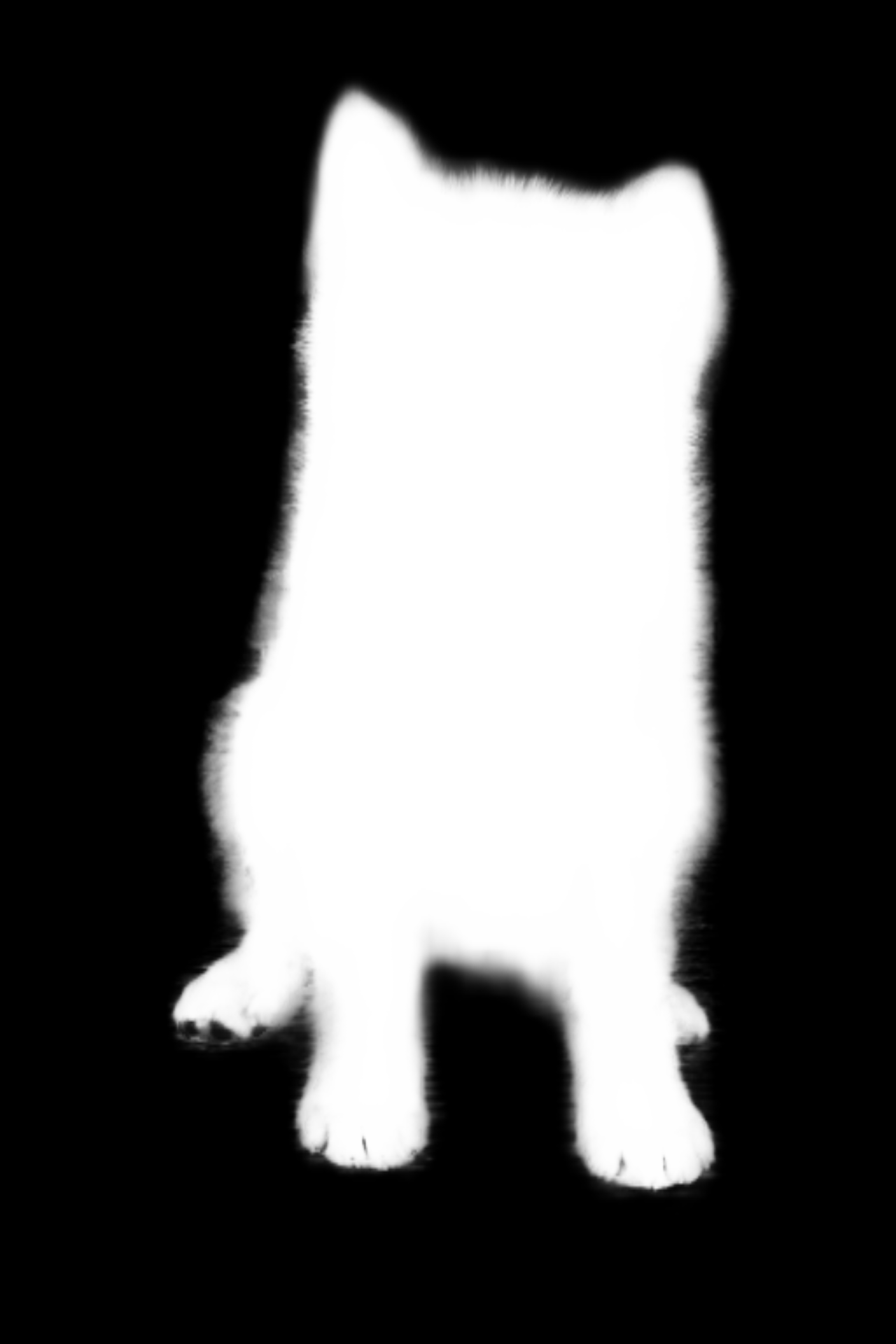}}
		\caption*{Ours}
	\end{subfigure}
	
	\caption{Results of samples from Adobe Composition-1k. Masks are generated by DeepLab v3+.}
	\label{fig:adobe_deeplab}
	\vspace{-3mm}
\end{figure*}

We further evaluate the proposed method on real images. We employ DeepLab v3+ \cite{chen2018encoder} as the segmentation method to demonstrate the performance of our method in practice. We only test on some images which have semantic objects that can be segmented by DeepLab v3+, and we adopt the segmentation as the input mask of our method. The results of potted plant images which are gathered from Google are shown in Figure \ref{fig:plant_deeplab}. The results of images from Supervise.ly Person dataset \cite{Supervise.ly} can be viewed in Figure \ref{fig:sup_deeplab}. In addition, Figure \ref{fig:adobe_deeplab} also demonstrate the results of samples from Adobe Composition-1k  \cite{xu2017deep}.

\section{Larger Unknown Area for Adobe DIM}
We also evaluate the performance of different methods with a larger unknown area in trimap for Adobe DIM \cite{xu2017deep}. We apply the dilation and erosion twice with a 25$ \times $25 kernel to generate a larger unknown area. Since errors are only computed in the unknown area, the quantitative results of our method and Guided Filter \cite{he2010guided} are also changed. The experiment results are shown in Table \ref{tab:ours} and Figure \ref{fig:test-data}.

\begin{table}[h]
	\small
	\centering
	\begin{tabular}{lcccc}  
		\toprule
		Methods & MSE & SAD & Grad. ($ \times 10^3 $) &Conn. ($ \times 10^3 $)\\
		\midrule
		Guided Filter & 0.112 & 24.49 & 21.33 & 14.25\\
		Adobe DIM & 0.146 & 21.36 & 20.54 & 14.20\\
		Our method & \textbf{0.017} & \textbf{3.29} & \textbf{8.32} & \textbf{1.89}\\        
		\bottomrule
	\end{tabular}
	\caption{The quantitative results with larger unknown area in trimap on the MAT-2793 testing set}
	\label{tab:ours}
\end{table}

\begin{figure*}[h]
	\foreach \t in {1,2,4,6}{%
		\centering
		\begin{subfigure}{0.3\columnwidth}	
			\centering
			{\includegraphics[width = 1\columnwidth]{figures/data/test_image_\t.jpg}}
		\end{subfigure}	
		\begin{subfigure}{0.3\columnwidth}
			\centering
			{\includegraphics[width = 1\columnwidth]{figures/data/test_alpha_\t.jpg}}
		\end{subfigure}	
		\begin{subfigure}{0.3\columnwidth}
			\centering
			{\includegraphics[width = 1\columnwidth]{figures/data/test_mask_\t.jpg}}
		\end{subfigure}
		\begin{subfigure}{0.3\columnwidth}
			\centering
			{\includegraphics[width = 1\columnwidth]{figures/supp/dim_trimap/trimap/\t.png}}
		\end{subfigure}
		\begin{subfigure}{0.3\columnwidth}
			\centering
			{\includegraphics[width = 1\columnwidth]{figures/supp/dim_trimap/alpha/\t.png}}
		\end{subfigure}
		\begin{subfigure}{0.3\columnwidth}
			\centering
			{\includegraphics[width = 1\columnwidth]{figures/data/test_idf_\t.jpg}}
		\end{subfigure}
		
	}
	\centering
	\begin{subfigure}{0.3\columnwidth}	
		\centering
		{\includegraphics[width = 1\columnwidth]{figures/data/test_image_10.jpg}}
		\caption*{Image}
	\end{subfigure}	
	\begin{subfigure}{0.3\columnwidth}
		\centering
		{\includegraphics[width = 1\columnwidth]{figures/data/test_alpha_10.jpg}}
		\caption*{Ground Truth}
	\end{subfigure}	
	\begin{subfigure}{0.3\columnwidth}
		\centering
		{\includegraphics[width = 1\columnwidth]{figures/data/test_mask_10.jpg}}
		\caption*{Mask}
	\end{subfigure}
	\begin{subfigure}{0.3\columnwidth}
		\centering
		{\includegraphics[width = 1\columnwidth]{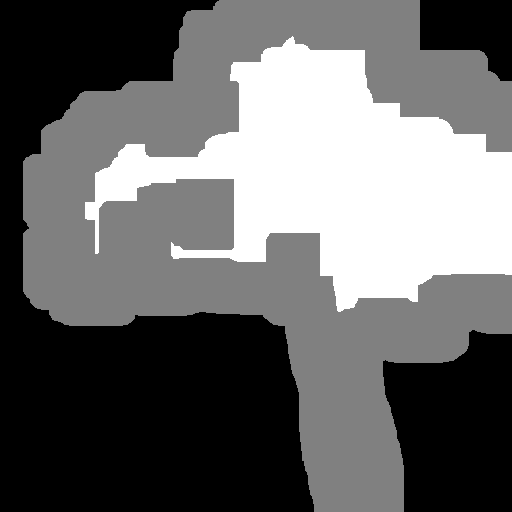}}
		\caption*{Trimap}
	\end{subfigure}
	\begin{subfigure}{0.3\columnwidth}
		\centering
		{\includegraphics[width = 1\columnwidth]{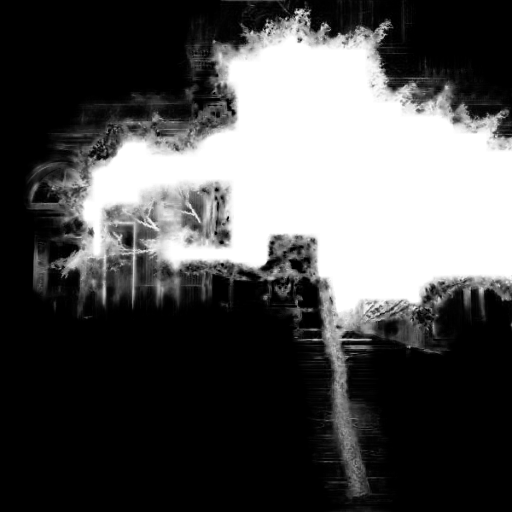}}
		\caption*{Adobe DIM}
	\end{subfigure}
	\begin{subfigure}{0.3\columnwidth}
		\centering
		{\includegraphics[width = 1\columnwidth]{figures/data/test_idf_10.jpg}}
		\caption*{Ours}
	\end{subfigure}
	
	\caption{The visual comparison results with larger unknown area in trimap on MAT-2793 testing set. The trimap is only for Adobe DIM.}
	\label{fig:test-data}
	\vspace{-3mm}
\end{figure*}

\end{appendices}

\end{document}